\documentclass[preprint,3p,number,12pt]{elsarticle}

\usepackage{amssymb}

\usepackage{multirow}
\usepackage{epstopdf}
\usepackage{color}
\usepackage [ruled, lined, linesnumbered] {algorithm2e}
\usepackage{amssymb, amsmath, amsthm}
\usepackage[figuresright]{rotating}

\newdefinition{defn}{Definition}

\journal{SCIENCE CHINA: Information Sciences}

\begin{document}

\begin{frontmatter}



\title{Matching Weak Informative Ontologies}

\author[seuadd]{Peng Wang}
\ead{pwang@seu.edu.cn}

\address[seuadd]{School of Computer Science and Engineering, Southeast University, Nanjing 210096, China}

\begin{abstract}

Most existing ontology matching methods utilize the literal information to discover alignments. 
However, some literal information in ontologies may be opaque and some ontologies may not have sufficient literal information.
In this paper, these ontologies are named as weak informative ontologies (WIOs) and it is challenging for existing methods to matching WIOs.   
On one hand, string-based and linguistic-based matching methods cannot work well for WIOs. 
On the other hand, some matching methods use external resources to improve their performance, but collecting and processing external resources is still time-consuming.  
To address this issue, this paper proposes a practical method for matching WIOs by employing the ontology structure information to discover alignments. 
First,  the semantic subgraphs are extracted from the ontology graph to capture the precise meanings of ontology elements. 
Then, a new similarity propagation model is designed for matching WIOs. 
Meanwhile, in order to avoid meaningless propagation, the similarity propagation is constrained by semantic subgraphs and other conditions. 
Consequently, the similarity propagation model ensures a balance between efficiency and quality during matching. 
Finally, the similarity propagation model uses a few credible alignments as seeds to find more alignments, and  some useful strategies  are adopted to improve the performance.
This matching method for WIOs has been implemented in the ontology matching system Lily. 
Experimental results on public OAEI benchmark datasets demonstrate that Lily significantly outperforms most of the state-of-the-art works in both WIO matching tasks and general ontology matching tasks.
In particular, Lily increases the recall by a large margin, while it still obtains high precision of matching results.

\end{abstract}


\end{frontmatter}

\section{Introduction}
More and more ontologies are created and used distributively by different communities in the past few decades. 
Ontology users or engineers would integrate or process multiple ontologies in practical applications. 
However, ontologies themselves could be heterogeneous. It is necessary to integrate various ontologies and enable cooperation between them. 
Ontology matching, which discovers alignments between ontologies, aims to provide a common layer from which heterogeneous ontologies could exchange information in semantically sound manners \cite{Doan02, Aumueller05}.

Many ontology matching methods have been proposed \cite{Rahm01, Kalfoglou03, Shvaiko05,   Shvaiko2012, Euzenat2013,  Cerdeira2015, Ochieng2018}. 
Generally, calculating the literal similarity is the most popular matching technique. 
However, not all ontologies provide sufficient, clear, and precise literal information for describing the semantics of elements in ontologies. 
For example, in the medical domain, the adult mouse anatomy ontology 
\footnote{http://web.informatik.uni-mannheim.de/oaei/anatomy11/index.html}
uses unique codes like MA\_0000436 to name concepts, 
and every term in geneontology
\footnote{http://geneontology.org/}
has a unique seven digit identifier GO ID like  GO:0006915.
In many ontologies, we also notice that some elements have no enough comments or labels to help users to understand their meanings. 
For another example, the 248-266 ontologies in the OAEI (Ontology Alignment Evaluation Initiative) \footnote{http://oaei.ontologymatching.org} benchmark dataset are such special cases, in which most labels are meaningless strings and even do not have any comment. 
Actually, these examples are not rare in real-world ontologies, and users still usually meet ontology elements with opaque labels or lack of sufficient information that cannot help users to understand the elements.
In some application domains such as semantic Web of Things \cite{Pfisterer2011}, electric power grid \cite{Tang_Grid19, Huang_JPES15} and  Industry 4.0 \cite{Bader_ESWC20}, ontologies often miss lexical layer instead of lots of human-unreadable IP addresses, linked sensors, electric switches, machine tools and products.
This paper calls an ontology without sufficient or clear literal information the weak informative ontology (WIO).

For this situation, string-based and linguistic-based matching methods cannot work well and will miss a lot of alignments, which leads to low recall of matching results. 
Therefore, it is necessary to find a feasible solution to solve the problem of matching weak informative ontologies. 
Utilizing the structure information is a natural idea to compensate for string-based and linguistic-based methods.  
Here, the structure-based ontology matching does not mean matching geometrical graphs, which cannot deal with the semantic information in ontologies. Namely, its matching results would have no meaning in semantics. 
Therefore, traditional graph matching algorithms \cite{Conte04} are not suitable here.
 
Most structure-based ontology matching methods \cite{Falcon2008, RiMOM09} usually employ the similarity propagation idea \cite{Melnik02, Jeh02}: \emph{similar objects are related to similar objects}, which assures that more alignments can be found by providing few alignments as seeds. 
Several similarity propagation models have been proposed for matching database schemas or XML schemas \cite{Melnik02, Jeh02, Blondel04, DoanA2005}. 
However, these algorithms cannot be used for ontology matching problem directly. 
For example,  Blondel's graph matching algorithm \cite{Blondel04}  cannot obtain good alignments for matching ontologies, because this algorithm directly treats an ontology as a graph and ignores the semantic information in the ontology.

To address the issue of matching WIOs, this paper attempts to propose a practical matching method which is inspired by  our previous work \cite{Wang09ASWC}. 
First, we generalize and redefine the problem as the weak informative ontology matching problem. 
Second, we introduce the semantic subgraph to describe elements in ontologies, which is the foundation of our solution, and present the algorithm for extracting semantic subgraphs. 
Third, we address how to use semantic subgraphs to build a matcher, which can provide credible seeds for subsequent similarity propagation. 
Fourth, we focus on discussing the new similarity propagation model for ontology matching and corresponding similarity propagation strategies. 
Finally, we provide more comprehensive experimental results to demonstrate the performance of our method.

The main original contributions of this paper are as follows.
(1) This paper proposes semantic subgraphs to capture precise meanings of ontology elements.
(2) A new similarity propagation model based on semantic subgraphs is proposed for matching weak informative ontologies. 
This is a novel similarity propagation model for matching ontologies. 
The propagation conditions are not only strict but also reasonable for ontology characteristics. 
In addition, the similarity propagation is constrained by semantic subgraphs, that can avoid meaningless propagation and improve the matching performance.
Therefore, compared to the classical general similarity propagation model Similarity Flooding \cite{Melnik02}, this model is more specific and ensures the balance between matching efficiency and quality. 
(3) Based on the semantic subgraphs, a matcher using semantic description documents is proposed, and it provides few credible alignments as seeds to the similarity propagation model. 
 Moreover, the similarity propagation model adopts some useful strategies to improve the matching performance. 
(4) Experimental results show that our method performs well not only for matching weak informative ontologies but also for matching general ontologies.
Especially, our method increases the recall by a large margin while still obtains high precision of matching results.

The remainder of this paper is organized as follows:
Section 2 addresses the weak informative ontology matching problem.
Section 3 presents an overview of the proposed method.
Section 4 discusses the ontology graph and its processing. 
Section 5 introduces the basic principle of semantic subgraphs.
Section 6 details the method for extracting semantic subgraphs from ontology graphs.
Section 7 discusses the similarity propagation model based on semantic subgraphs.
Section 8 analyzes and compares different propagation scale strategies.
Section 9 describes the matcher using semantic subgraphs.
Section 10 presents the experimental results and corresponding discussions.
Section 11 is an overview of related work and Section 12 is the conclusion.

\section{Problem Statement and Analysis}
Usually, an ontology contains concepts, properties, instances and axioms. 
We follow the work in \cite{Shvaiko2012, Euzenat2013} and give a formal definition for the ontology matching problem. 

\begin{defn} 
[\emph{Ontology Matching}]
The matching between two ontologies $O_{1}$ and $O_{2}$ is:
$\mathcal {M} \!=\!\{m_{k}|m_{k}\!=<\!\!e_{i},e_{j},r,s\!\!>\}$,
where $\mathcal {M}$ is an alignment; 
$m_{k}$ denotes a correspondence with a tuple $<e_{i},e_{j},r,s>$; $e_{i}$ and $e_{j}$ represent the expressions which are composed of elements from $O_{1}$ and $O_{2}$ respectively; 
$r$ is the semantic relation between  $e_{i}$ and $e_{j}$, and $r$ could be equivalence {($=$)}, generic/specific ($\sqsupseteq$/$\sqsubseteq$), disjoint ($\perp$) and overlap ($\sqcap$), etc.;  
$s$ is the confidence about an alignment and typically in $[0,1]$ range. 
Therefore, an alignment $\mathcal {M}$ is a set of correspondences $m_{k}$. 
\end{defn}

This paper only focuses on correspondences of $concept-concept$ and $property-property$ with equivalence  relation.

\begin{figure*}[htb!]
	\begin{centering}
	\includegraphics[trim = 0mm 0mm 0mm 0mm, clip,  width=1.00\textwidth]{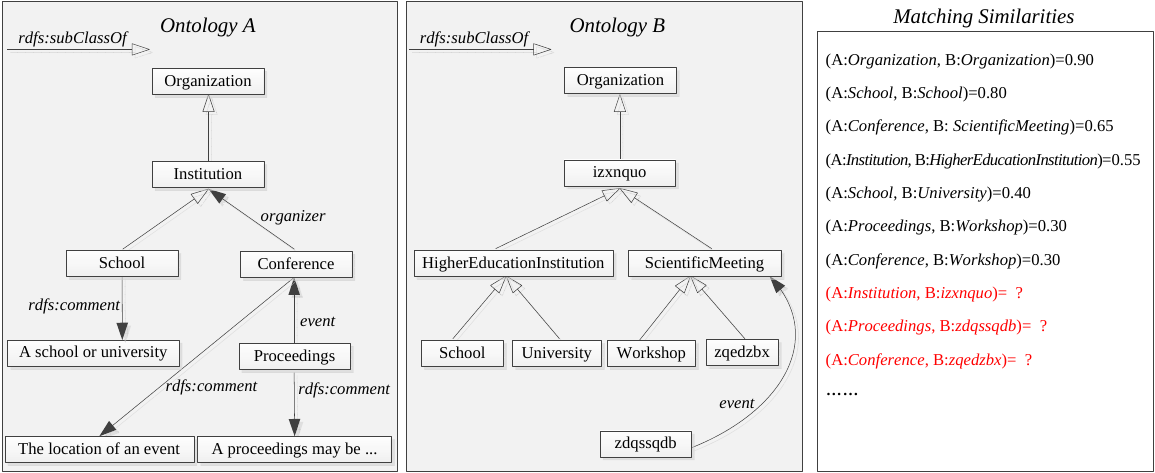}
	\caption{An weak informative ontology matching scenario}
	\label{wio_example}
	\end{centering}
\end{figure*}

If we have enough and clear literal information about all elements, alignments can be discovered easily. 
However, real-world ontologies cannot always provide such ideal literal information. 
Figure~\ref{wio_example} shows a matching scenario in OAEI. 
All concepts of ontology $A$ have necessary comments or labels. However, in ontology $B$, some elements, such as $izxnquo$ and $zdqssqdb$, have meaningless labels and there are no comments to explain these concepts. 
In this matching scenario, for some elements like \emph{A:Conference} and \emph{B:ScientificMeeting}, we can easily calculate matching similarities by linguistic-based or string-based techniques. 
However, we cannot calculate the similarities for element pairs which contain opaque elements. 
For example, the similarity between \emph{A:Institution} and \emph{B:izxnquo} is difficult to be calculated.
Such opaque elements appear in ontologies for two possible reasons. 
First, ontology engineers do not provide sufficient annotations or comments for all elements, that will make some elements lack enough information to be understood. 
Second, to name elements, ontology engineers would use some special codes, which cannot be understood by others. 
For instance, concept $Address$ may be named as $Add$ , $Adr$ or $Dizhi$ (in Chinese spelling). We call such elements the weak informative elements and define as follows.

\begin{defn} 
[\emph{Weak Informative Element}]
Given an ontology element $e$ and a lexicon $\mathcal {L}$ , 
let $D_e$ be the set of words in $e$'s literal information including identities, labels, comments and annotations. 
$|D_e \cap \mathcal {L} |$ denotes the number of words contained in both $\mathcal {L}$ and $D_e$. 
$D_e$ can not be $\emptyset$ and $|D_e| \geq 1$. 
If $\frac{|D_e \cap \mathcal {L} |}{|D_e|}<\phi$, $e$ is a weak informative element, where $\phi$ is a small threshold value in [0,0.5]. 
\end{defn}

In this paper,  $\phi$ is set to 0.25, and the lexicon $\mathcal {L}$ is WordNet \footnote{https://wordnet.princeton.edu}. For a weak informative element $e$, its meaning is difficult to be understood according to corresponding literal information. 

Therefore, if an ontology contains a certain proportion weak informative elements, it is difficult to be understood. We call such an ontology the weak informative ontology and define as follows.

\begin{defn} 
[\emph{Weak Informative Ontology}]
Given an ontology $O$ with $N$ elements, let $w$ be the number of weak informative elements and $\delta$ be a predefined threshold in $[0,1]$, if $\frac{w}{N}>\delta$, $O$ is a weak informative ontology (WIO).  
\end{defn}

For the reason that building an ontology is a time-consuming and error-prone process, an ontology would contain some weak informative elements, which make matching methods difficult to find alignments.  
Table~\ref{wio_survey} shows a survey of weak informative ontologies for 10 ontologies in OAEI2008 benchmark dataset \footnote{http://oaei.ontologymatching.org/2008/benchmarks/}. 
We manually examine these ontologies and count the number of weak informative elements in each ontology. 
Column 2 to column 4 are the ratio of weak informative elements in concepts, properties and instances, respectively. 
The $w/N$ value is the total ratio between number of weak informative elements and total number of elements.
The predefined threshold $\delta=0.25$, that means if the total ratio is larger than 0.25, users cannot understand the ontology well. 
Consequently, 7 ontologies in Table~\ref{wio_survey} are weak informative ontologies. 
The last column shows matching results ( average, best and worst F1-measure) of  6 classic systems \cite{OAEI08} (Aflood, AROMA, ASMOV, DSSim, GeroMe and MapPSO) and 2 new systems (AML \cite{AML2013, AML2018} and LogMap \cite{LogMap2011, LogMap2018}). 
According to matching results, there is a correlation between whether the ontologies are WIO and the average matching results of ontology matching system over them. 

\begin{table*}
	\caption{A survey of weak informative ontologies on the sample of OAEI2008 benchmark ($\delta$=0.25)}
	\begin{center}
		\begin{tabular}{lllllll}
			\hline
			\noalign{\smallskip} 
			ontology & \multicolumn{3}{c}{weak informative elements/all elements}  
			& \multirow{2}{*}{$w/N$} & \multirow{2}{*}{is WIO} & matching results
			\\ 
			 \cline{2-4}
			ID & concepts & properties & instances & & & (Average, Best, Worst )\\
			
			\noalign{\smallskip}
			\hline
			101 & 0/34     & 0/72       & 0/55      & 0.00 & No     & 0.97, \quad  1.00, \quad 0.88                      \\ 
			201 & 3/34     & 11/72      & 0/55      & 0.07 & No     & 0.81, \quad  1.00, \quad 0.12                      \\ 
			202 & 31/34    & 65/72      & 55/55     & \textbf{0.94} & \textbf{Yes}    & 0.46, \quad  0.88, \quad   0.05                \\ 
			203 & 0/34     & 5/72       & 15/55     & 0.12 & No     & 0.97, \quad 1.00, \quad 0.88                     \\ 
			248 & 31/34    & 65/72      & 22/55     & \textbf{0.75} & \textbf{Yes}    & 0.34, \quad   0.81, \quad  0.04                  \\ 
			250 & 31/34    & 0/8        & 22/55     & \textbf{0.55} & \textbf{Yes}    & 0.34, \quad   0.53, \quad   0.08                \\ 
			252 & 24/27    & 63/73      & 21/55     & \textbf{0.70} & \textbf{Yes}    & 0.33, \quad  0.82, \quad    0.06         \\ 
			254 & 33/36    & 0/8        & 23/55     & \textbf{0.57} & \textbf{Yes}    & 0.28, \quad    0.43, \quad 0.00                \\ 
			258 & 31/34    & 64/72      & 0/0       & \textbf{0.90} & \textbf{Yes}    & 0.20, \quad    0.62, \quad 0.02                \\ 
			260 & 29/32    & 0/9        & 23/55     & \textbf{0.54} & \textbf{Yes}    & 0.35,  \quad   0.57, \quad 0.03                  \\
		\hline
		\end{tabular}
	\end{center}
	\label{wio_survey}
\end{table*}

Therefore, it is necessary to find a new way to discover alignments for weak informative ontologies. 
Researchers have to utilize ontology structure information to compensate for the lack of literal information. 
Although an ontology can be represented as a graph, ontology matching is not equal to the graph matching problem, 
in which a correspondence between two elements not only means that the elements are similar in geometrical perspective, more importantly, but also are similar in semantics.
Moreover, graph matching is an NP problem \cite{Conte04}, so it cannot match ontologies efficiently. 
For example, we attempt to use the graph matching API provided by $SOQA-SimPack$ \cite{SOQA06} to match ontology graphs,  and we find that it needs more than several days or even several weeks for matching two normal size ontologies. 
Most importantly, graph topology information cannot represent semantics in ontologies, so a geometrical graph similarity cannot imply the semantic similarity. 
Therefore, it is not suitable to treat ontology matching as the graph matching problem.

Most structure-based ontology matching methods are inspired by the simple idea: \emph{similar objects are related to similar objects}. This idea also derives some heuristic rules, such as \emph{concepts may be similar when their super/sub concepts are similar} and \emph{concepts may be similar when they have similar instances}. 
These rules have been used by some matching systems \cite{QOM04, Noy03}. 
However, if ontologies only have some opaque literal information, namely the $\delta$ is high, heuristic rules usually cannot work. 
The reason is that the similarity between elements' neighbors cannot be determined without enough clear literal information, so it causes that the similarity between elements cannot be determined too.

A reasonable solution for this problem is similarity propagation, namely, the similarity between elements can propagate to their neighbors in the graph, then similarity propagation can produce more and more similarities.
After each propagation, all similarities are normalized. The propagation process is terminated until the similarities are converged. Based on such similarity propagation idea, several similarity propagation models have been proposed \cite{Jeh02, Blondel04, HU05, Leicht06, Melnik02}. 
Among these models, similarity flooding \cite{Melnik02} is the most classical one. 
The similarity flooding includes three steps:
(1) constructing pairwise connectivity graph;
(2) constructing induced propagation graph;
(3) computing fixpoint values for matching.
Similarity flooding is a versatile matching algorithm and can be implemented easily, but it is not sensitive to the initial similarity seeds. 
It means that different initial seeds would produce similar matching results.  Similarity flooding algorithm has been used for schema matching in database and XML data \cite{Rahm01}. However, similarity flooding is not a perfect algorithm. Melnik and his colleagues summarized six disadvantages \cite {Melnik02}, for example,  having similar neighbors is the necessary precondition of this algorithm. 
After we try to use similarity flooding to match ontologies directly, we also find the algorithm cannot work effectively for matching WIOs. 
First, similarity flooding does not consider the similarities between edges, which are properties in ontologies, so the property similarities between ontologies cannot be calculated. 
Secondly, the maximum pairwise connectivity graph is $N_{A}\!*\!N_{B}$ ($N_{A}$ and $N_{B}$ are the numbers of edges in two ontologies, respectively), and it will greatly increase the time complexity for fixpoint computing and space complexity for storing the pairwise connectivity graphs.  
Moreover, in real-world matching tasks, the ontology graph may have thousands edges, so the corresponding pairwise connectivity graphs would become too large to be handled. 
For above reasons, the similarity flooding algorithm cannot be used directly for matching WIOs.
This paper aims to modify the similarity flooding and proposes a new propagation model  to solve the matching problem for WIOs.

\section{Overview of the Methodology }

\begin{figure*}[b]
	\begin{centering}
		\includegraphics[trim = 0mm 0mm 0mm 0mm, clip,  width=1.00\textwidth]{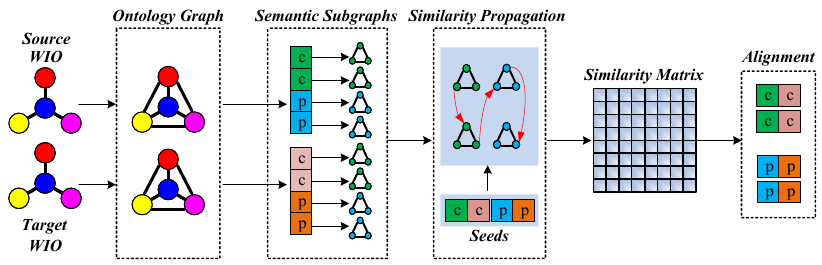}
		\caption{Overview of matching weak informative ontologies}
		\label{overview_wio}
	\end{centering}

\end{figure*}

Figure~\ref{overview_wio} depicts an overview of the proposed method for matching weak informative ontologies, which involves three steps: (1) building the ontology graph from the WIO, (2) extracting semantic subgraphs from the ontology graph,  and (3) calculating similarity propagation to obtain similarity matrix and the alignment. 

We first use the hybrid ontology graph to represent the WIO for distinguishing multiple properties between concepts,  then explicitly describe the containers and collections in the ontology graph, afterwards, enrich the ontology by discovering hidden semantics, furthermore, refine the ontology graph by removing annotation and definition triples. 
As a result, according to the original source WIO and target WIO, we build two ontology graphs, which can clearly describe the semantic information in ontologies.

For each concept (denoted by c in Figure~\ref{overview_wio} ) or property (denoted by p in Figure~\ref{overview_wio} ) in the ontology graph, we extract the corresponding semantic subgraph, which can precisely describe the meaning of the concept or property. 
This step is based on the commonsense that people can understand a concept or property with limited semantic information,  rather than the whole ontology. 
Specially, in this paper, we apply a circuit model to efficiently rank triples and then extract semantic subgraphs. 
More concretely, in the circuit model, the conductivity simulates the capability of conveying information, the voltage indicates the capability of preserving information, and the current denotes the semantic information flows on edges in the ontology graph. 

Based on semantic subgraphs, this paper proposes a novel similarity propagation model for matching the weak informative ontologies. 
Considering the characteristics of ontologies, we design a strong constraint condition during similarity propagation, which not only avoids the performance drawbacks of similarity flood model but also can handle the correspondences between properties in ontology matching. 
Additionally, the propagation model employs the updating mechanism, credible seeds, penalty, termination condition, and propagation scale strategies in order to ensure a balance between matching efficiency and quality.
In particular, the initial credible seeds in propagation are provided by a matcher based on semantic subgraphs,  i.e. the matcher calculates the similarities between the semantic description documents, which are constructed from semantic subgraphs.
Lastly, after the similarity propagation, we obtain the similarity matrix and then extract the alignment from it. 


\section{Ontology Graph}


An ontology is composed of statements, which are triples like $<s,p,o>$. $s$, $p$ and $o$ stand for the subject, predicate and object in a statement, respectively. 
There are three kinds of ontology resources: URIs resources, literals and blank nodes. 
In a triple, the subject can be URIs resources or blank nodes but not literals,  and the predicate must be URIs resource. 
Let $sub(O)$, $pred(O)$ and $obj(O)$ represent the sets of ontology resources for subject, predicate and object, respectively. 
An ontology can be directly converted into a raw ontology graph.

\begin{defn} 
[\emph{Raw Ontology Graph}]
An ontology $O$ can be represented by a labeled directed graph $G_r=<V, E, l_V, l_E>$, where $V$ and $E$ refer to sets of vertices and edges, respectively. 
$V =\{x| x\in sub(O) \cup obj(O)\}$, $E=\{y|y\in pred(O)\}$, $l_V$ and $l_E$ are functions which map vertices and edges to their labels. 
Two vertices and an edge strictly correspond to a triple in an ontology. 
$G_r$ is called the raw ontology graph.
\end{defn} 

\begin{figure*}[htb!]
	\begin{centering}
		\includegraphics[trim = 0mm 0mm 0mm 0mm, clip,  width=1.00\textwidth]{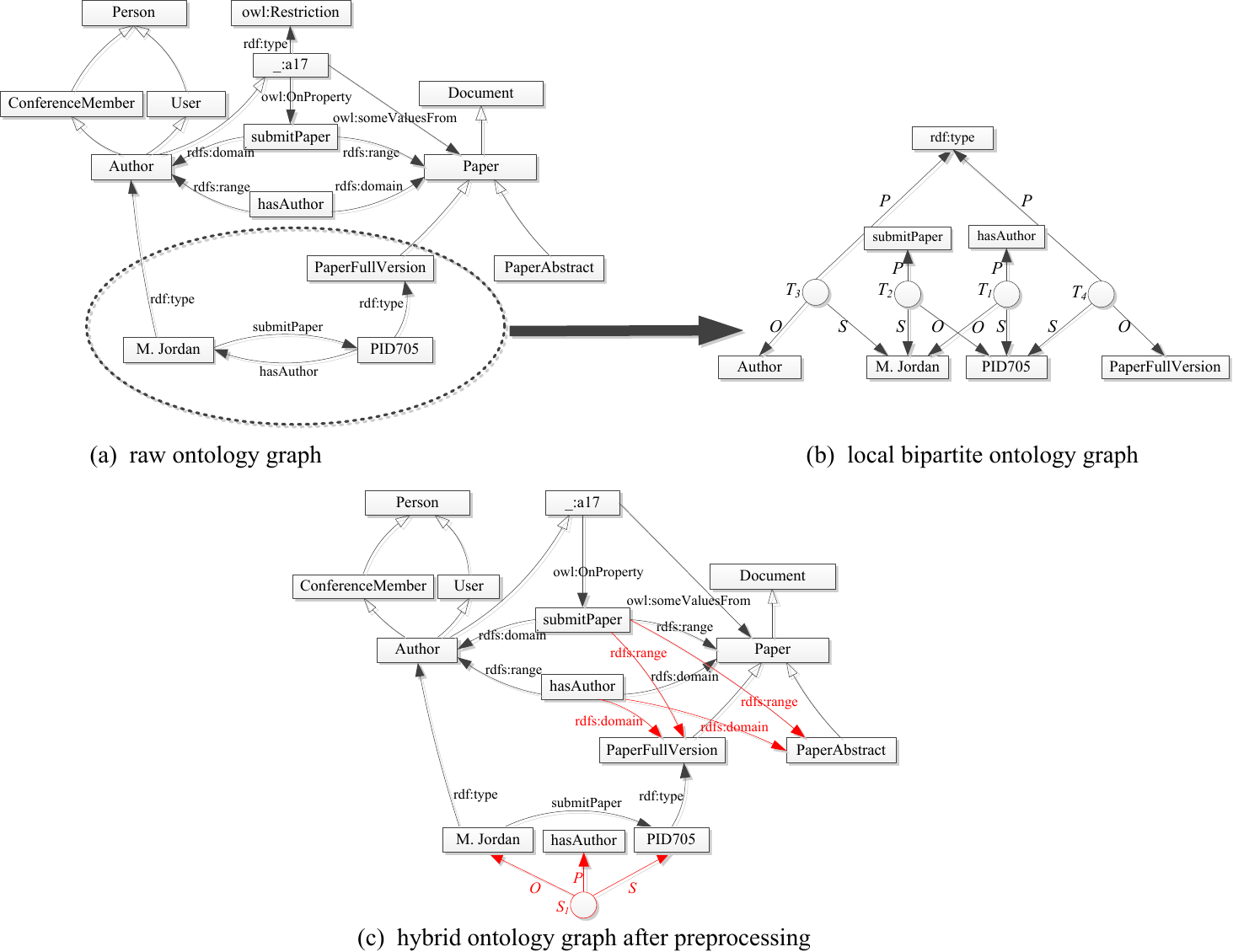}
		\caption{Ontology graph processing}
		\label{ontgraph_proc}
	\end{centering}
\end{figure*}

Figure~\ref{ontgraph_proc} (a) shows a simple raw ontology graph for describing conference knowledge. 
The hollow arrows represent \emph{rdfs:subClassOf}. The vertex \emph{\_a:17} denotes a blank node.

A raw ontology graph $G_r$ is a multigraph, in which more than one edge can exist between two vertices. 
A raw ontology graph and its adjacency matrix have two shortcomings. 
First, a property can appear at a vertex and an edge at the same time, but the adjacency matrix cannot represent this fact. 
Second, the adjacency matrix cannot distinguish multiple edges between two vertices in the multigraph. 
To deal with the first shortcoming, we record all statements about properties with extra space. 
The second shortcoming can be solved by the bipartite graph, which converts each triple $<s,p,o>$ into three triples: $e_s=<T_i,S,s>$, $e_p=<T_i,P,p>$ and $e_o=<T_i,O,o>$, where $T_i$ assures the three triples can be reconverted into the original triple. 
As shown in Figure~\ref{ontgraph_proc}(b),  multiple edges in part of Figure~\ref{ontgraph_proc}(a) are represented as the bipartite graph.

Although the bipartite graph can distinguish multiple edges between two vertices and then all ontology resources can appear at vertices, it still has disadvantages: 
(1) A bipartite graph size is three times larger than the original raw ontology graph. 
(2) The bipartite graph cannot directly describe semantic relations between elements, and that makes it difficult to analyze the ontology graph.
To avoid these disadvantages, we present the hybrid ontology graph, which combines advantages of the raw ontology graph and the bipartite graph. 
In a hybrid ontology graph, if two vertices have only one edge, the triple can be represented in the raw ontology graph, and if there are more than one edges between two vertices, we convert some edges into a bipartite graph.

\begin{defn} 
[\emph{Hybrid Ontology Graph}]
Given a raw ontology graph $G_r=<V, E, l_V, l_E>$, let $E_p(v_m,v_n)= \{<v_m,p_i,v_n> \in E\}$ be all edges from $v_m$ to $v_n$.  $|E_p(v_m,v_n)|$ is the number of edges between $v_m$ and $v_n$. The hybrid ontology graph for $G_r$ is $G_h=<V', E', l_V, l_E>$, which can be constructed according to following rules:
\begin{enumerate} [(1)]
\item If $|E_p(v_m,v_n)|=1$, the edges and vertices are directly converted into $G_h$.
\item If $|E_p(v_m,v_n)|>1$, then randomly convert $|E_p(v_m,v_n)|-1$ edges into a bipartite graph and add them into $G_h$.
\end{enumerate}
\end{defn} 

A hybrid ontology graph can be implemented by an adjacency matrix. 
Edges using the bipartite graph style can be reconverted into the original triples. 
The corresponding hybrid ontology graph of Figure~\ref{ontgraph_proc}(a) is shown in Figure~\ref{ontgraph_proc}(c), in which an edge is represented by bipartite graph.


There is some semantic information hidden in the ontology which would be useful for understanding the ontology, meanwhile, some triples are not useful for describing the semantics of elements. Therefore, a hybrid ontology graph needs to be further processed, that includes processing containers and collections, enriching and refining the ontology graph.

\textbf{Phase 1: Processing containers and collections}

In RDF language\footnote{https://www.w3.org/RDF/}, containers and collections are used to describe sets of resources.
Containers are represented in \emph{rdf:Bag}, \emph{rdf:Seq} and \emph{rdf:Alt}.
Collections are represented in \emph{rdf:List}.
Although container and collection simplify the ontology, but their semantics cannot be clearly represented in ontology graphs.
Hence, we use  semantically equivalent statements to replace  containers and collections.
For example,  in a triple $<Beatles, artist, \_a11>$, $\_a11$ is a \emph{rdf:Bag} blank node  and has four members: \emph{John, Paul, George, Ringo}, 
then we use four triples to replace the original description as:  $<Beatles, artist,  John>$,   $<Beatles, artist, Paul>$, $<Beatles, artist, George>$, $<Beatles, artist, Ringo>$.

\textbf{Phase 2: Enriching the ontology graph}

The enriching process discovers more hidden semantics and represents them clearly.
In fact, although discovering hidden semantics can be seen as ontology reasoning, 
we use enriching rules instead of existing reasoners because the rules have more flexible operations on ontologies and are  independent of ontology languages. 
\begin{enumerate} [Step 1.]
\item Enriching domain and range: In the property hierarchy, the domain and range of a super property can be inherited by its sub properties. Such semantics are clearly declared in ontology graphs.

\item Enriching concept axioms: Concept axioms include \emph{owl:oneOf}, \emph{owl:intersectionOf}, \emph{owl:unionOf}, \emph{owl:equivalentClass}, etc. 
For example, if an \emph{owl:intersectionOf} axiom defines a complex concept $A \sqcap B$ which has a sub concept $C$, then $A \sqsupset C$ and $B \sqsupset C$  are added into the ontology graph.

\item Enriching property axioms: Property axioms include \emph{owl:SymmetricProperty}, \linebreak \emph{owl:TransitiveProperty}, \emph{owl:equivalentProperty}, etc. 
For example, an axiom \linebreak \emph{owl:SymmerticProperty} declares predicate $p$ is symmetric, and there is a triple $<s,p,o>$ ($s$ is subject and $o$ is object), then a new triple $<o,p,s>$ is added into the ontology graph.

\item Enriching \emph{owl:sameAs} axioms: If some resources are declared by owl:sameAs, they share same semantic information.

\item Enriching properties in the concept hierarchy: The property of a super concept can be inherited by its sub concepts. For example, given $<p, \emph{rdfs:domain}, A>$ and $<B, \emph{rdfs:subClassOf}, A>$, then $<p, \emph{rdfs:domain}, B>$ holds.
\end{enumerate}

\textbf{Phase 3: Refining the ontology graph}

In an ontology graph, annotation and definition triples are removed, that make relations between elements clearer.
\begin{enumerate} [Step 1.]
\item Removing  annotations: Remove triples including \emph{rdfs:label}, \emph{rdfs:comment}, \emph{rdfs:seeAlso}, \emph{rdfs:isDefineBy} and \emph{owl:AnnotationProperty}.

\item Removing the ontology head information: The ontology head information, which is between two $<$\emph{owl:Ontology}$>$ tags, only describes the information about the ontology. So it can be removed.

\item Removing the version information: We remove the version information such as \emph{owl: versionInfo}, \emph{owl: backwardCompatibleWith}, \emph{owl:incompatibleWith}, \emph{owl:priorVersion}, \emph{owl:DeprecatedClass} and \emph{owl: DeprecatedProperty}.

\item Removing \emph{rdf:type} statements pointing to the metadata: For example, $<$\emph{Game}, \emph{rdf:type}, \emph{owl:Class}$>$ should be removed, but $<StarWars, \emph{rdf:type}, Game>$ should be kept.

\item Removing \emph{owl:Thing} and \emph{owl:Nothing}: They are the global concept $\top$ and the empty concept $\bot$, which do not describe the semantic relations between elements.
\end{enumerate}

In Figure~\ref{ontgraph_proc}(c), red edges are added after ontology graph processing. 

Table~\ref{change_og} shows changes of graph size in ontology graph processing.
 Here, we use six datasets in OAEI2007 \footnote{http://oaei.ontologymatching.org/2007/}: 
 (1) 101, 301, 302, 303 and 304 in the benchmark task, 
 (2) \emph{Cmt} and \emph{Edas} in the conference task, 
 (3) \emph{source} and \emph{target} in the directory task, 
 (4) \emph{mouse} and \emph{nci} in the anatomy task, 
 (5) AGOVOC, NALT and GEMET in the food task and (6) Brinkman and GTT in the library task. 
 Column 2 to 5 show changes of ontology graph size during different  processing phases. 
 When an ontology has containers and collections, phase 1 adds new triples and deletes old triples for the containers or collections, so the graph size change is uncertain. 
 The enriching phase usually increases the graph size. 
 Many hidden semantic information was discovered in the benchmark. 
 The refining phase reduces the ontology graph size, especially in the last 4 datasets that have a lot of annotation information.

\begin{table}
	\caption{Changes of ontology graph size in ontology graph processing}
	\label{change_og}
	\begin{center}
	\begin{tabular} {llllll}
			\hline
			\noalign{\smallskip} 
			& $G_r$ & phase1 & phase2 & phase3 & $G_h$	\\
			\noalign{\smallskip}
			\hline

			benchmark	&3772		&3734		&6086		&4409 	&4453 \\
			conference	&2133		&2133		&2774		&2436		&2436 \\
			directory	&18970	&18970	&18970	&9483		&9483 \\
			anatomy	&51268	&51268	&51268	&28367	&28367 \\
			food		&337037	&337037	&337037	&93312	&93312 \\
			library	&161806	&161806	&161806	&20156	&20156 \\

		\hline
		\end{tabular}
	\end{center}
\end{table}

\section{Semantic Subgraph}

The semantic subgraph, which is the foundation of our solution for matching weak informative ontologies, is used for precisely describing the meaning for each ontology element.
An ontology is composed of elements including concepts, properties, and instances. 
Understanding ontology elements is one of the most frequent operations in ontology applications such as semantic searching and ontology matching. 
A single element is meaningless without the semantic context, which contains relative statements or triples in an ontology. 
The meaning of an element in an ontology can be described by relative statements or triples. 
However, there is no way to determine which statement should be selected and whether the selected statements are enough. 
For the former question, people often intuitively think statements that directly contain the element are what we need, but it is not always correct. 
For example, given an element \emph{MichaelJordan}, if we know
$<\emph{MichaelJordan}, playIn, Bulls>$,
the direct statement $<\emph{MichaelJordan}, drive, car>$  conveys less information about \emph{MichaelJordan} than the indirect statement $<Bulls, isA, \emph{NBATeam}>$.  
In order to answer the latter question, we must find a way to measure whether we have gathered enough semantic contexts.
The semantic context of an ontology element is composed of some relevant statements (or triples), which is a subgraph in the ontology graph. Such subgraphs are called semantic subgraphs of ontology elements.

\begin{defn} 
	[\emph{Semantic Subgraph}]
	Given an  element $e$ in a hybrid ontology graph $G_h$, its semantic subgraph $G_s(e)$ is composed of \emph{top-k} $(top\mbox{-}k\in \mathbb{N})$ related triples that describe $e$. 
	 $G_s(e)\subseteq G_h$  and $G_s(e)$ has following features:
\begin{enumerate}
	\item \emph{The size of $G_s(e)$  is limited.} We believe that only  \emph{top-k} related triples can accurately describe the context of an element, namely, the semantic interpretation of an element does not need all knowledge in the ontology.
	\item \emph{$G_s(e)$ does not emphasize semantic completeness.} A semantic subgraph collect as much information about the element as possible until it can distinguish an element from other elements.
	\item \emph{$G_s(e)$ is unique.} Two elements with different semantics have different semantic subgraphs.
	\item \emph{$G_s(e)$ prefers triples related to $e$.} Different triples have different capabilities for describing an element. 
	Ranking all triple about $e$  according to related scores from high to low, $G_s(e)$ fist selects higher triples. 
	\item \emph{Closer triples do not mean more related to $e$.} It is possible that triples far from the element may be more important than closer ones.
\end{enumerate}
\end{defn}

These five features assure that a semantic subgraph provides a clear, accurate and credible semantic description for an ontology element.
Concept and property are the two important basic types of elements in an ontology. 
We will discuss the method for extracting semantic subgraphs for given concepts or properties from ontology graphs.

\section{Extracting Semantic Subgraphs based on Circuit Model}
According to the principle of semantic subgraphs, in order to extract the semantic subgraph for a given element, 
we first rank all related triples in ontology graphs, then select  \emph{top-k} triples to compose the semantic subgraph.
This paper proposes a circuit model to efficiently rank triples and extract semantic subgraphs. 
In this section, we first introduce the circuit model, then calculate  conductivity, finally discuss the extraction algorithm.

\subsection{Circuit model}

In order to extract semantic subgraphs, this paper utilizes circuit to model the semantic information propagation in an ontology.
First, assuming that semantic information about $s$ in an ontology graph is a measurable value in [0, 1], and it comes from vertex $s$ with initial semantic information value 1,  then flows to other vertices $s_i$ through some triples. 
Since these triples have resistance to semantic information flow, the semantic information reaching $s_i$ will have some losses. Such information about $s$ will flow in the ontology graph continually. 
The semantic subgraph of $s$ is composed of paths which start from $s$ and have more semantic information about $s$.
This process corresponds to an electrical circuit model. 
The semantic information from $s$ corresponds to adding +1 volt on $s$. 
The semantic information on a path from $s$ to $s_i$ corresponds to the current reaching $s_i$ through this path. 
The resistance in information flow corresponds to the electric resistance.

Similar circuit model has been used by Faloutsos et al \cite{Faloutsos2004} for discovering connection subgraphs in social networks. 
Based on this previous work, this paper proposes a modified model based on characteristics of ontologies for extracting semantic subgraphs.

In the circuit model, the capability of conveying information is the conductivity $C$, the capability of keeping information is the voltage $V$, and current $I$ denotes the total information flows on edge per unit time. 

A connection subgraph connects the source vertex $s$ and the target vertex $t$ \cite{Faloutsos2004}. 
However, a semantic subgraph  only has the source vertex $s$ (the given element). 
Therefore, the first modification is adding a sink node $z$ as the target vertex into the ontology graph, in which each vertex has an edge to $z$. 

Given two vertices $u$ and $v$, let $I(u,v)$ denote the current from $u$ to $v$, $V(u)$ and $V(v)$ be the voltages on $u$ and $v$, 
and $C(u,v)$ and $R(u,v)$ be the conductivity and resistance on the edge between $u$ and $v$,  $C(u,v)=1/R(u,v)$. 
Then an ontology graph is converted into a circuit, which has the following initial conditions:
\begin{equation}
V(s)=1, V(z)=0
\end{equation}

All voltages and currents in the circuit can be computed according to Ohm's law and Kirchhoff's law.

The conductivity from any vertex to sink node $z$ is:
\begin{equation}
C(u,z)=\lambda \sum \limits_{w \neq s} C(u,w)
\end{equation}
Here, $\lambda$ denotes the current loss coefficient, and  $1\geq \lambda>0$. 

\begin{defn} 
[\emph{Delivered Current}]
The delivered current $\hat{I}(P)$  in a prefix-path $P=(s=u_1,\ldots,u_i)$ is the volume of electrons that arrives at $u_i$  through $P$.
\end{defn}

The delivered current can be calculated by:
\begin{equation}
\hat{I}(s=u_1,...,u_i)=
\hat{I}(s=u_1,...,u_{i-1})
\frac{I(u_{i-1},u_i)}
{I_{out}(u_{i-1})}
\end{equation}
 $I_{out}(u)$ is the total current come from $u$.

In physics, the delivered current describes the remaining current through a path from $s$. 
Here it denotes the amount of semantic information about $s$ in a path. 
In other words, the delivered current measures the relevance between a path and $s$, that further means that triples in the path are relevant to $s$. 
Therefore, a subgraph about $s$ can be regarded as the combination of some  prefix-paths.
The captured flow of a subgraph can be defined as follows. 

\begin{defn}
[\emph{Captured Flow}]
The captured flow of subgraph $G_s$ is the sum of all the delivered current in the prefix-path in $G_s$:
\begin{equation}
CF({G_s}) =\sum\limits_{P = (s,...,t) \in {G_s}} {\hat I(P)}
\end{equation}
\end{defn}

Therefore, for all subgraphs having \emph{k} triples, the subgraph with the maximum capture flow is the semantic subgraph. 
In other words, a semantic subgraph is determined by the captured flow on paths, which contain relevant triples about $s$. 
A subgraph with more captured flow has more information about $s$.
In the above process, we do not rank the triples directly, but use the maximum captured flow to indirectly realize the ranking. 
Namely, a new triple is added in a semantic subgraph is always the one that can increase maximum captured flow. 

Extracting semantic subgraph can be divided into two sub problems: 
(1) Traversing all prefix-paths from $s$ to $z$ and calculating their delivered currents. 
(2) Searching all $k$-size subgraphs combined by prefix-paths and calculating their capture flows. 
\textbf{The subgraph with the maximum captured flow is the semantic subgraph.} 
Howerver, the two sub problems are NP problems.
Faloutsos et al. proposed a greedy algorithm called \emph{DisplayGeneration} \cite{Faloutsos2004} to efficiently discover connection subgraphs. 
The greedy idea prefers prefix-paths bringing the maximum fraction between delivery current and new nodes. 
Our algorithm of extracting semantic subgraphs is also based on this greedy algorithm. 

The time complexity of solving the circuit model is $O(|V|^3+|E|\times m\times k)$. 
The first part is the complexity of solving the circuit linear equation, and the later part is the complexity of extracting the semantic subgraphs by the greedy algorithm. 
$|V|$ is the number of vertex in ontology graph, $|E|$ is the number of edges, $m$ is the maximum length of the paths from $s$ to $z$ and $m<|V|$.
$k$ is the size of the semantic subgraph. 
Usually, $k$ is a constant and is far smaller than $|V|$. 
Most ontology graphs are sparse , in which $|E|$ and $|V|$ are linear relationship. 
Thus the later part is $O(|V|^2)$ approximately. 
Then the performance is dominated by the time for solving the circuit linear equation.
Therefore, the approximate time complexity is $O(|V|^3)$

The space complexity of the circuit model is $O(|V|^2+|V|m)$, where the former part is the space for solving the linear equation, and the later part is the space for the greedy algorithm. The total space complexity can also be simplified to $O(|V|^2)$.

\subsection{Conductivity calculation}

If the conductivity between vertices was 1, which means that the whole information is delivered. 
However, in an ontology graph, the semantic information would have losses when it flows through triples, so the edge conductivity should be a value in [0,1]. 
Based on some ontology characteristics, we first derive some heuristic rules to measure capability of delivering information, then calculate the conductivity. 

\textbf{Rule 1: Frequency rule}


Here we first introduce an attenuation function:
\begin{equation}
g(x,m) = {1 \over 2}({1 \over x} + (1 - {{\log x} \over {\log (m + \varepsilon )}})){\rm{  }},m \ge x \ge 1
\end{equation}
where $x$ and $m$  are  variables,  and $\varepsilon$ is a small positive constant to assure $log(m+ \varepsilon)>0$. 
Comparing with $f(x)=1/x$, $g(x,m)$ can slowly decrease when $x$ increases.

Let $f(e)$ be the number of triples in an ontology graph $G_h$ in which $e$ appears.
The weight of $e$ is:
\begin{equation}
{\mu _f}(e) = g(f(e),max_{e_i \in G_h} f(e_i))
\end{equation}

The frequency rule applies to concept, property and metadata, and it means more frequently used elements deliver less information. 

\textbf{Rule 2: Hierarchy rule}

Hierarchy is the important way to organize concepts and properties. 
The higher level elements have less capability of delivering information. 
It can be measured by following weight.
\begin{equation}
{\mu _H}({e}) = {{dh({e})} \over {max_{e_i \in G_h} dh(e_i)}}
\end{equation}
$dh(e)$ denotes the depth of the element in the hierarchy.
The hierarchy rule applies for concept and property.

\textbf{Rule 3: Instance space rule} 

An instance space in an interpretation function on an ontology $O$. 
Given a concept $C$, its instance space is the set ${I_{sp}}(C) = \{ a| <a,\emph{rdf:type},C> \in O\}$. 
Given a property $P$, its instance space is ${I_{sp}}(P) = \{  <a,b>|<a,P,b> \in O , {a} \in Dom(P),{b} \in Rng(P)\} $, 
where $Dom(P)$ and $Rng(P)$ are the domain and range of $P$, respectively.


A larger instance space indicates that a concept has more instances, then the concept has a higher possibility of being on the top level, which means that it will deliver less information. 
Similarly, if a property instance space is larger, then the property is used more frequently, so it conveys less information. 
Such a weight can be calculated as:
\begin{equation}
{\mu _{Isp}}(e) = g(|{I_{sp}}(e)|,max_{e_i \in G_h} |I_{sp}(e_i)|)
\end{equation}
where $|I_{sp}(e)|$ is the size of the instance space of $e$. 

\textbf{Rule 4: Instance property description rule}
This rule is based on the assumption that more key instances would have more relative triples.
This assumption is based on our experience that key instances in an ontology are described in more details than satellite instances.
Therefore, this assumption usually holds. 
Given an instance $a$ and its relative triples like $<a,p_m,b>$, the number of $p_m$ can be used to measure the importance of the instance. The formula to calculate the weight is:
\begin{equation}
{\mu _{Id}(a) =
{ 
	{dp(a) + op(a)} 
	\over 
    { max_{a_i\in G_h} dp(a_i) + max_{a_i\in G_h} op(a_i) }
}
}
\end{equation}
where $dp(a)$ denotes the number of \emph{DatatypeProperty} and $op(a)$ denotes the number of \emph{ObjectProperty}.

\textbf{Rule 5: Few instance rule}

Let $a$ be an instance of concept $C$. 
Concepts having few instances will have more information. 
The weight can be calculated as:
\begin{equation}
{\mu _{Io}}({a}) = g(|C({a})|,max_{a_i \in G_h}|C({a_i})|)
\end{equation}
where $|C(a)|$ is the number of instances of $C$.

Based on the above five rules, the weights for concept $C$, property $P$, instance $I$ and metadata $M$ are as follows:
\begin{eqnarray}
\mu (C) &=& {\gamma _{C1}} \times {\mu _f}(C) + {\gamma _{C2}} \times {\mu _H}(C) 
 {}+ {\gamma _{C3}} \times {\mu _{Isp}}{\rm{(C)}} \\
\mu (P) &=& {\gamma _{P1}} \times {\mu _f}(P) + {\gamma _{P2}} \times {\mu _H}(P) 
{}+ {\gamma _{P3}} \times {\mu _{Isp}}(P) \\
\mu (I) &=& {\gamma _{I1}} \times {\mu _{Id}}(I) + {\gamma _{I2}} \times {\mu _{Io}}(I) \\
\mu (M) &=& {\mu _f}(M)
\end{eqnarray}
where ${\gamma _{C1}} + {\gamma _{C2}} + {\gamma _{C3}} = 1$, ${\gamma _{P1}} + {\gamma _{P2}} + {\gamma _{P3}} = 1$, and ${\gamma _{I1}} + {\gamma _{I2}} = 1$.

Finally, the conductivity for any triple $t=<s,p,o>$ can be obtained based on the weights of $s$, $p$ and $o$. Since $s$ and $o$ are relevant to other triples, their weights should be divided by the degrees.
\begin{equation}
w(t) = {{{{\mu (s)} \over {\deg ree(s)}} + \mu (p) + {{\mu (o)} \over {\deg ree(o)}}} \over 3}
\end{equation}

Although five rules are intuitive and empirical, they essentially conform to entropy in information theory, namely, lower possibility events deliver more information.

\subsection{Semantic subgraph extraction algorithm}

\begin {algorithm}[htb!]
\DontPrintSemicolon

\KwIn{ontology graph $G$}
\KwOut{$S$:semantic subgraphs for elements}
\BlankLine

\Begin{
	$S \leftarrow \varnothing$\;
	\tcp*[l]{calculate weights}
	$G_w \leftarrow GetGraphTripleWeight(G)$ \;
	\tcp*[l]{traverse all elements}
	\ForEach {$e_i \in G$}{
		\tcp*[l]{add 1 volt}
		$SetVolt(e_i) \leftarrow 1$ \;
		\tcp*[l]{solve circuit equation}
		$SolveLinearSystem(G_w, e_i)$ \;
		\tcp*[l]{extract a semantic subgraph}
		$G_s(e_i) \leftarrow Extract(G_w,e_i,N)$ \;
		$S \leftarrow S \cup G_s(e_i)$ \;
	}
}
\BlankLine
\caption{Extracting semantic subgraphs}
\end{algorithm}

\begin{figure*}[htb!]
		\begin{centering}
		\includegraphics[trim = 0mm 0mm 0mm 0mm, clip,  width=1.00\textwidth]{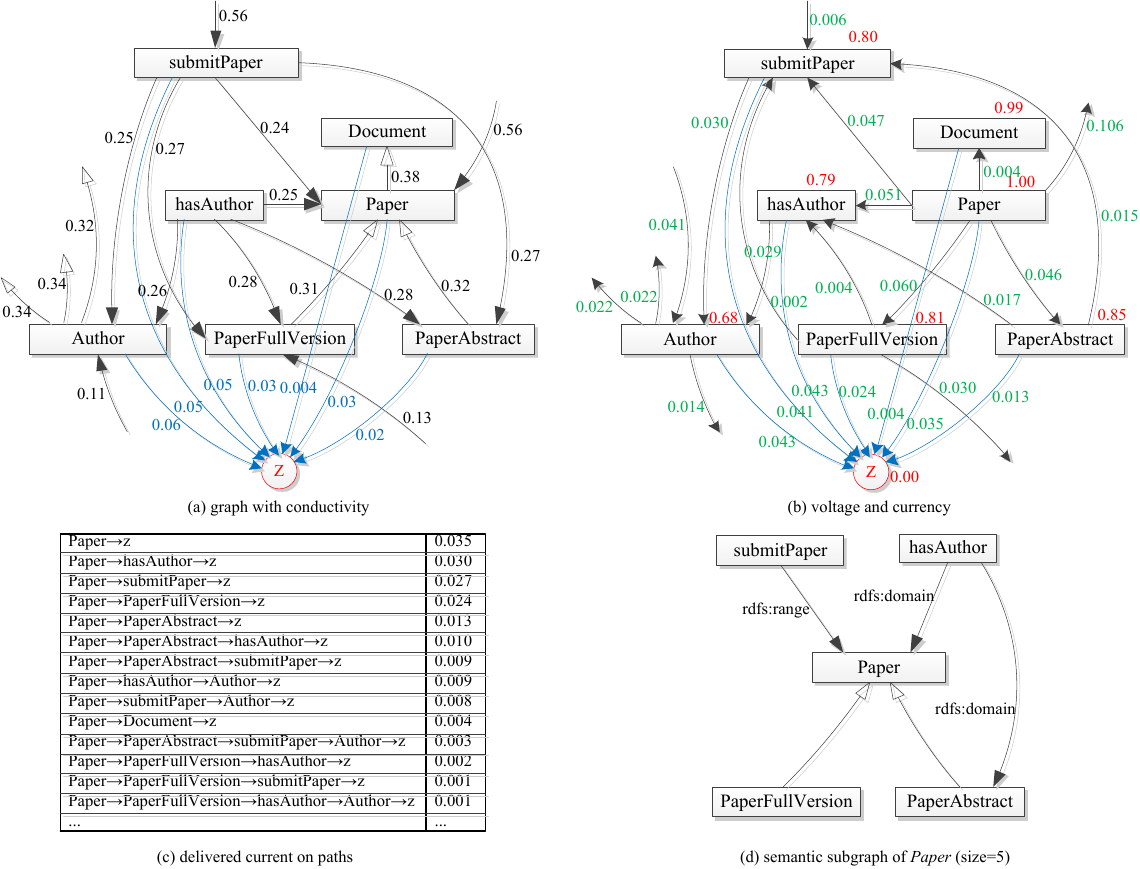}
		\caption{An example of extracting the semantic subgraph for a concept \emph{Paper}}
		\label{sgextract}
	\end{centering}
\end{figure*}

After calculating the conductivity, the circuit model can be used to extract semantic subgraphs for concepts and properties. 
Algorithm 1 describes the extraction process. 
A concept always locates at a vertex in an ontology graph. 
First, for any concept $e_i$, 1 volt is added to the vertex (line 5). 
Then, the circuit equation is solved (line 6). 
Third, according to greedy algorithm \emph{DisplayGeneration} \cite{Faloutsos2004}, the $k$-size subgraph  with the maximum captured flow is the semantic subgraph (line 7). 
Finally, semantic subgraph $G_s(e_i)$ is obtained.
Note that, since a property can appear on a vertex and an edge at the same time, the extraction for a property has little different to a concept.  
For a property $P_i$, if 1 volt was only added to $P_i$, the triple $<c_j,P_i,o_j>$ would have less current, but it is useful for describing the semantic of $P_i$. 
Therefore,  an edge from $P_i$ to $c_j$ is added to overcome this problem,  that will increase the current on edge $<c_j,P_i,o_j>$.


Figure~\ref{sgextract} shows the process of extracting the semantic subgraph for concept \emph{Paper} in Figure~\ref{ontgraph_proc}. 
In Figure~\ref{sgextract}(a), sink node $z$ is added and all conductivities are calculated. 
In Figure~\ref{sgextract}(b), +1 voltage is added on $Paper$ and $z$ has 0 voltage, then all voltages and currents are computed. 
Red values are voltages and green values are currents. 
Figure~\ref{sgextract}(c) shows some paths from $Paper$ to $z$ and their delivered currents. 
All paths are ranked by the delivered currents. 
In order to extract a 5-size semantic subgraph for $Paper$, we can add these paths one by one according to their delivered currents from high to low, until the semantic subgraph has 5 triples. 
Such subgraph is the semantic subgraph of $Paper$.
The 5 triples in this semantic subgraph are more relevant to $Paper$ than the ones in paths with lower delivered currents.
The captured flow of this semantic subgraph is maximum in all subgraphs with 5 triples.

\section{Similarity Propagation Model Based on Semantic Subgraphs}
Based on  semantic subgraphs, we propose a novel similarity propagation model for matching weak informative ontologies. 
We first discuss the similarity propagation condition, then present the detail of the propagation model including the updating, seeds, penalty, and termination condition.  

\subsection{ Similarity propagation condition}

Ontology graph consists of triples like $<\!\!s_{i},p_{i},o_{i}\!\!>$.
In ontology matching, a reasonable similarity propagation should consider both vertices ($s_i$ and $o_i$) and edges ($p_i$).
Similarity flooding model \cite{Melnik02} presumes that each edge pair ($p_x,p_y$) has 1.0 similarity value,
and the similarity of vertex pair ($s_x,s_y$) will be propagated to another vertex pair ($o_x,o_y$).
This propagation condition has three disadvantages:
(1) It would produce a large number of alignment candidates and generate a large scale pairwise connectivity graph;
(2) It would produce many incorrect alignment candidates.
(3) It cannot deal with the correspondences between properties in ontology matching.

In order to avoid these disadvantages, this paper proposes a new propagation condition for ontology matching, namely, the strong constraint condition (SC-condition).

\begin{defn}
[\emph{Strong Constraint Condition}]
Given two triples $t_{i}\!=<\!\!s_{i},p_{i},o_{i}\!\!>$ and $t_{j}\!=<\!\!s_{j},p_{j},o_{j}\!\!>$,
and let $S_{s}$, $S_{p}$ and $S_{o}$ denote the corresponding similarities of $(s_i,s_j)$, $(p_i,p_j)$ and $(o_i,o_j)$, respectively.
Similarities can be propagated only $t_{i}$ and $t_{j}$ satisfy following three conditions:
\begin{enumerate}[(1)]
\item
In $S_{s}$, $S_{p}$ and $S_{o}$, at least two similarities must be larger than threshold $\theta$;
\item
If $t_{i}$ includes ontology language primitives, the corresponding positions of $t_{j}$ must be same primitives;
\item
$t_{i}$ or $t_{j}$ has at most one ontology language primitive.
\end{enumerate}
\end{defn}

Condition (1) ensures that the final similarity results are creditable after propagating. 
The ontology language primitives refer to RDF vocabularies and OWL vocabularies\footnote{https://www.w3.org/TR/owl-guide/}.
Condition (2) ensures that two triples use same ontology language primitive to describe semantics.
For example, $<\!\!Conference\_Paper,rdfs\!\!:\!\!subClassOf,Paper\!\!>$ and $<\!\!Paper,rdfs\!\!:\!\!subClassOf,Document\!\!>$ use
the RDF primitive $rdfs\!\!:\!\!subClassOf$ as predicate, so the similarities can be propagated between them.
Condition (3) ensures that there is no ontology definition and declaration triples during propagating,
because such triples may cause incorrect matching results.
For example, without condition (3), two triples $<\!\!PhDStu, rdf\!\!:\!\!type, rdfs\!\!:\!\!Class\!\!>$ and $<\!\!Paper, rdf\!\!:\!\!type, rdfs\!\!:\!\!Class\!\!>$ will
cause wrong alignment: $PhDStu\!=\!Paper$.

After one propagation, the similarity of an element pair will be increased by the sum of
other two pairs. Taking the similarity $S_s$ as an example after $i^{th}$ propagation, its new similarity is:
\begin {equation}
S^{i}_{s}=S^{i-1}_{s}+w_{po}\times S^{i-1}_{p}\times S^{i-1}_{o}
\end {equation}
Analogously, the $S^i_p$ and $S^i_o$ are:
\begin {equation}
S^{i}_{p}=S^{i-1}_{p}+w_{so}\times S^{i-1}_{s}\times S^{i-1}_{o}
\end {equation}
\begin {equation}
S^{i}_{o}=S^{i-1}_{o}+w_{sp}\times S^{i-1}_{s}\times S^{i-1}_{p}
\end {equation}
$w_{po},w_{so}$ and $w_{sp}$ are propagation factors, which will be discussed latter. In addition, all similarities will be normalized after each  propagation.

\begin{figure*}[htb!]
	\begin{centering}
	\includegraphics[trim = 0mm 0mm 0mm 0mm, clip, width=1.00\textwidth]{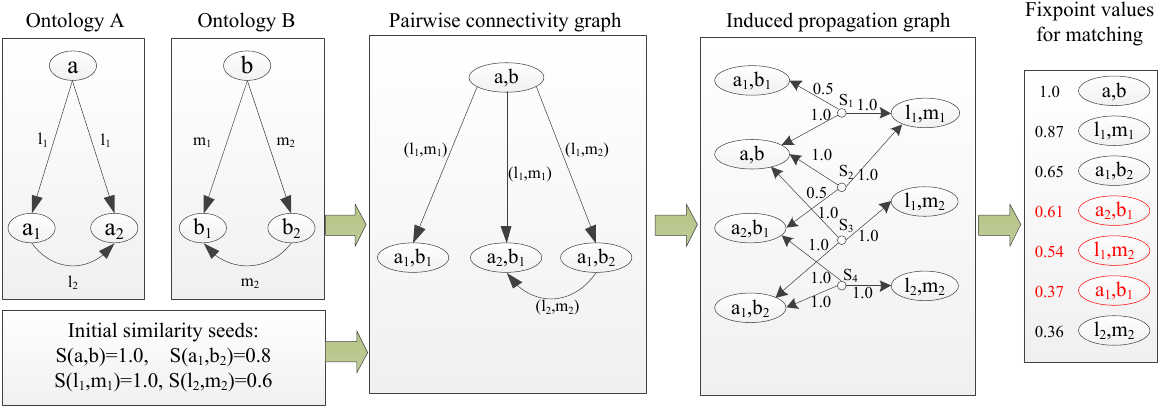}
	\caption{Similarity propagation model with SC-condition}
	\label{sp_sc}
	\end{centering}
\end{figure*}

\subsection{ Similarity propagation model}

Our new similarity propagation model contains three steps. 
Given two ontology graphs and initial similarity seeds, we first construct pairwise connectivity graph, 
then get induced propagation graph, and finally obtain the new similarities by fixpoint value calculation.
Figure~\ref{sp_sc}  illustrates the similarity propagation model with three steps as follows.

\textbf{Step1: Constructing pairwise connectivity graph}

Traditional similarity flooding model is not sensitive to initial similarity seeds, so all initial similarity values can be set to 1.0. 
However, in ontology matching, similarity propagation cannot use the same setting, because it
will not only cause very large pairwise connectivity graph but also generate many wrong correspondences.
In our view, the quality of the initial similarity seeds is very important for matching ontologies, namely, credible correspondences would generate more credible correspondences during similarity propagation. 
Wrong correspondences are noise in similarity propagation. 
Therefore, this paper tries to use few high quality correspondences as initial similarity seeds, which can be calculated by string-based matching methods or provided manually.

According to initial similarity seeds and the SC-condition, the pairwise connectivity graph can be constructed as shown in  Figure~\ref{sp_sc}. 
The pairwise connectivity graph is influenced by similarity seeds. 
Different seeds will cause different pairwise connectivity graphs.

\textbf{Step2: Constructing induced propagation graph}

According to formula (16)-(18),  similarities from two element pairs are propagated to the third pair.
The propagation factor measures how much similarity value can be propagated.
There are three kinds of propagation factor:
$w_{sp}$, $w_{so}$ and $w_{po}$. 

Take $w_{sp}$ as an example,
it denotes how much similarity value that comes from $S_s$ and $S_p$ can be propagated to $S_o$. Let $f_{sp}$
denote the number of the triple pairs having $(s_i,s_j)\xrightarrow[]{(p_i ,p_j )}(o_x,o_y)$ style in pairwise connectivity graph, then $w_{sp}=1/f_{sp}$.
$w_{so}$ and $w_{po}$ can be  calculated analogously.

The induced propagation graph can be represented by a bipartite graph as shown in Figure~\ref{sp_sc}, 
where $S_1$, $S_2$, $S_3$ and $S_4$ in induced propagation graph denote the four triples in pairwise connectivity graph.
Therefore, the similarities of properties in ontologies can also be propagated.
The weights of edges denote the propagation factors.
In the implementation, for the reason that propagation factors can be directly obtained
according to the pairwise connectivity graph, we record the propagation factors but need not to store the induced propagation graph.

\textbf{Step3: Computing fixpoint values}

The similarity propagation between ontology graphs can be computed iteratively until the final similarity matrix is converged. 
Under the SC-condition, the fixpoint values can be computed by formula (19), where normalization is omitted for clarity.

Actually, formula (19) is the synthesized style for formula (16)-(18).  
For each element pair $(x,y)$, which would be subject pair, predicate pair or object pair, its new similarity in the $(i+1)^{th}$ propagation contains four parts: 
\begin{enumerate}[(1)]
\item  The similarity in $i^{th}$ propagation; 
\item  The propagation similarity when $(x,y)$ is object pair;
\item  The propagation similarity when $(x,y)$ is subject pair; 
\item  The propagation similarity when $(x,y)$ is predicate pair.
\end{enumerate}

\begin {eqnarray}
s^{i + 1} (x,y) = && s^i (x,y) \nonumber\\
&& +\sum_{\substack{< a_u ,p_u ,x >  \in A \\ < b_u ,q_u ,y >  \in B}}
                                        s^i (a_u ,b_u )\cdot s^i (p_u ,q_u )\cdot w_{sp} \nonumber\\
&& + \sum_{\substack{< x,p_v ,a_v  >  \in A \\ <y,q_v ,b_v  >  \in B}}
                                        s^i (a_v ,b_v )\cdot s^i (p_v ,q_v )\cdot w_{po}{\;} \nonumber\\
&& + \sum_{\substack{< a_t ,x,c_t   >  \in A \\ <b_t ,y,d_t   >  \in B}}
                                        s^i (a_t ,b_t )\cdot s^i (c_t ,d_t ) \cdot w_{so}{\;\;\;\;} \nonumber\\
\end {eqnarray}

\subsection{Incremental updating for pairwise connectivity graph}

The SC-condition greatly reduces the scale of pairwise connectivity graph. 
After one similarity propagation, the similarity matrix will change, and new similarity values between elements are obtained. 
In Figure~\ref{sp_sc}, the red pairs are new similarities after one propagation. 
Therefore, for the next propagation, we need to construct a new pairwise connectivity graph. 
However, constructing a pairwise connectivity graph is a time-consuming process, because we need to check all similarity values and select right triples from the ontology graph. 

To reduce the  constructing cost, we adopt an incremental updating way. 
After one similarity propagation, the new pairwise connectivity graph dose not need to be reconstructed entirely, but can be extended based on the previous one. 
We only update the parts in the pairwise connectivity graph whose similarities have been changed. 
If an element pair has been in the pairwise connectivity graph and its similarity is smaller than $\theta$, we remove that element pair from the pairwise connectivity graph. 
If a new element pair has been discovered and its similarity is bigger than $\theta$, we add that element pair into the pairwise connectivity graph.
If two new triples satisfy the SC-condition, we add them to the new pairwise connectivity graph. Otherwise, we remove the triples that do not satisfy the SC-condition.

\begin{figure*}[htb!]
	\begin{centering}
	\includegraphics[trim = 0mm 0mm 0mm 0mm, clip, width=1.00\textwidth]{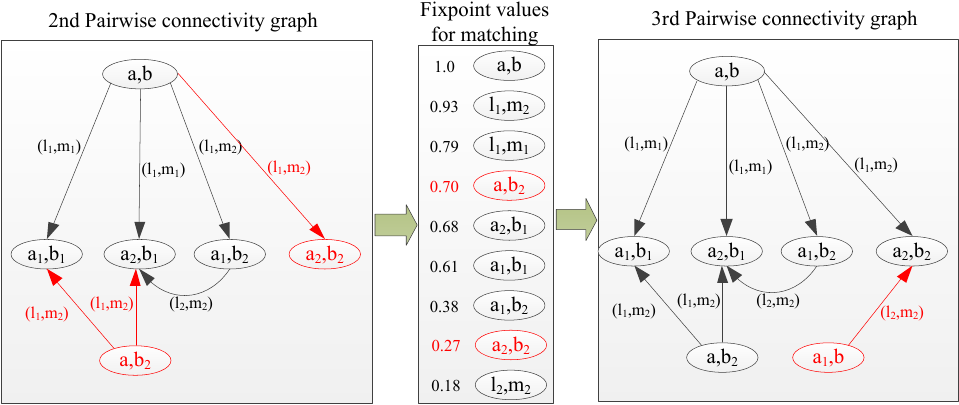}
	\caption{Updating pairwise connectivity graph}
	\label{up_pcg}	
	\end{centering}
\end{figure*}

Figure~\ref{up_pcg} illustrates the incremental updating for pairwise connectivity graphs. 
The second pairwise connectivity graph is constructed based on the first one in Figure~\ref{sp_sc}. 
The third connectivity graph is also constructed based on the second one. 
The red vertices and edges are new parts in pairwise connectivity graphs. 
New element pairs are also shown in red in the fixpoint values list. 

\subsection{Credible seeds}

For initial similarity seeds, we regard correspondences having high similarity value as credible seeds. 
We will keep these credible correspondences during the similarity propagation. 
This strategy has two advantages. 
First, it assures that some correct correspondences cannot be changed or be affected by other triples during similarity propagation. 
Second, it can avoid some unnecessary similarity propagation calculation.
 
If $S(a_i,b_j)$ is a credible seed, then all similarity propagation calculations like $S(a_i,b_x)$ and $S(a_y,b_j)$ can be skipped.
Hence, credible seeds can not only reduce the propagation cost, but also decrease the negative effect in propagation.

\subsection{Penalty in propagation}

For an ideal similarity matrix, correct correspondences should have higher confidence values and incorrect correspondences should have lower confidence values. 
However, the real-world similarity matrix is far from perfect.
Therefore, it is necessary to penalize the correspondences to improve propagation results . 
The penalty will make little influence for correspondences having high confidence value but reduce the potential wrong correspondences having lower confidence value.

We provide two penalty factors $p_a$ and $p_b$ as follows.
\begin {equation}
p_a=\frac{s(a_i,b_j)}{max(s_{max}(a_i,b_x),s_{max}(a_y,b_j))}
\end {equation}
 $s_{max}(a_i,b_x)$ is the maximum value in $i$-th row, $s_{max}(a_y,b_j)$ is the maximum value in $j$-th column. 
 Therefore, $p_a$ measures the ratio of similarity value $s(a_i,b_j)$ to the maximum value in its row and column.
\begin {equation}
p_b=\frac{1}{1+e^{-\alpha t}}, \mbox{ } t=(\frac{N+1}{n_i+1}/log(N+1)), \mbox{ } \alpha \geq 1
\end {equation}
 $N$ is the number of columns and rows in similarity matrix. 
 $n_i$ is the number of correspondences whose confidence values are larger than 0 in $i$-th column and $j$-th row. 
 We set $\alpha =3$ in the implementation.
 
After being penalized, the new similarity value is:
\begin {equation}
S^{'}(a_i,b_j)=S(a_i,b_j)\cdot p_a \cdot p_b
\end {equation}
$p_a$ penalizes the correspondences having low similarity values, and $p_b$ penalizes correspondences whose
column and row have too many correspondences with $S(x,y)\!>\!0$. 

\subsection{Termination condition}

Our similarity propagation model should satisfy three termination conditions: 
(1) The matrix norm between two sequential similarity matrices is not bigger than a given threshold. 
Propagation should assure that the final similarity matrix is convergent. 
Fortunately, Melnik and his colleagues have proved that fixpoint computing can be convergent if the pairwise connectivity graph is a strongly connected graph \cite{Melnik02}. 
(2) There is no update for the pairwise connectivity graph. 
(3) In a similarity propagation, to avoid the matrix needs too many times propagation to be convergent, we set the maximum propagation times as 8 in the implementation.


\section{Propagation Scale Choosing Strategies}

To improve the efficiency of similarity propagation and the quality of matching results, we design five propagation scale strategies to study what kind of graphs should be used in the similarity propagation. 
These strategies are about choosing the right parts in the ontology graph for similarity propagation. 

\begin{figure*}[htb!]
	\begin{centering}
	\includegraphics[trim = 0mm 0mm 0mm 0mm, clip, width=0.90\textwidth]{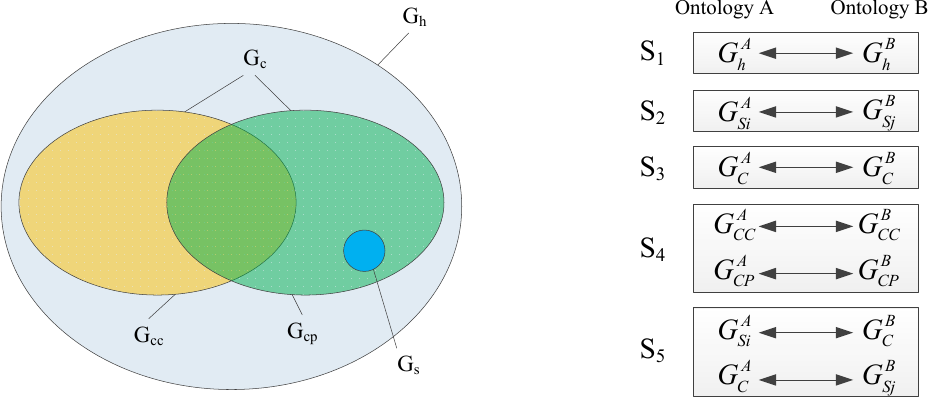}
	\caption{Five propagation scale strategies}
	\label{ps_strategy}
	\end{centering}
\end{figure*}

An element $e$ has a semantic subgraph $G_s(e)$. All semantic subgraphs of concepts are combined to a graph $G_{CC}$. All semantic subgraphs of properties are combined to a graph $G_{CP}$.  $G_{CC}$ and $G_{CP}$ may be overlapping. Let $G_C$ be the graph combined by all semantic subgraphs of concepts and properties. 
Then $G_C=G_{CC} \cup G_{CP}$. 
Meanwhile, $G_C$ is part of hybrid ontology graph $G_h$. 
The left of Figure~\ref{ps_strategy} illustrates the intersection relationships of these graphs.
According to these graphs, we design five propagation scale strategies as shown in the right of Figure~\ref{ps_strategy}. 

\textbf{Strategy 1: Full ontology graph propagation}

In similarity propagation, a simple and direct way is using full ontology graph as the propagation scale. 
For two ontologies $O_A$ and $O_B$, the similarity propagation is between two hybrid ontology graph $G_h^A$ and $G_h^B$.
Although there is no ontology information to be missed in this strategy, it also has two drawbacks: 
(1) During similarity propagation, the pairwise connectivity graph may become too large, then it will influence the efficiency of propagation and be difficult to be handled. 
Especially for the large scale ontology graph, it is possible to generate very large pairwise connectivity graph.
(2) More triples do not mean better propagation results. In full ontology graph, some triples are not important for describing semantics. 
In addition, too many triples may increase more uncertainty or noise in propagation and bring negative affection for matching results.

\textbf{Strategy 2: Independent semantic subgraph propagation}

A semantic subgraph is used to precisely describe an element. 
Therefore, if we constrain the propagation scale in semantic subgraphs, the propagation can avoid triples that is irrelevant to the element.
Given two elements $e_i$ and $e_j$ and corresponding semantic subgraphs $G_{Si}^A$ and $G_{Sj}^B$, the similarity
$S(a,b)$ is obtained by the similarity propagating between $G_{Si}^A$ and $G_{Sj}^B$. 
If two ontologies have $n$ and $m$ elements respectively, this strategy needs $n\!\times \!m$ times similarity propagations. For the reason that each semantic subgraph is small, the similarity propagation can be calculated quickly.

\textbf{Strategy 3. One combined semantic subgraph propagation}

We notice that different semantic subgraphs may be overlapping. 
In  strategy 2, some triples would be used in multiple similarity propagations. 
Therefore, we can combine all semantic subgraphs as the propagation scale. 
According to features of semantic subgraphs, the combined graph only  contains triples relevant to elements. 
It not only improves the propagation efficiency, but also removes the irrelevant information to avoid introducing propagation noise. 
Let $G_C^A$ and  $G_C^B$ be the combined semantic subgraphs of  $O_A$ and $O_B$, respectively, the similarity propagation is between  $G_C^A$ and  $G_C^B$.

\textbf{Strategy 4: Two separate combined semantic subgraphs propagation}

Strategy 3 can be redivided into a more concrete strategy. 
Since a $G_C$ is combined by a $G_{CC}$ and a $G_{CP}$,  the similarity propagation between concepts can be constrained in  $G_{CC}$, and the similarity propagation between properties can also be constrained in $G_{CP}$. 
Therefore, for ontology  $O_A$ and $O_B$, the similarities between concepts are calculated by the similarity propagation between  $G_{CC}^A$ and $G_{CC}^B$, and the similarities between properties can be calculated by the similarity propagation between $G_{CP}^A$ and $G_{CP}^B$.
From the ontology matching perspective, such propagation strategy will produce two similarity matrices: one is similarity matrix for concepts, another is similarity matrix for properties.

\textbf{Strategy 5: Hybrid semantic subgraph propagation}

In strategy 2, the similarity propagation is between two small semantic subgraphs. 
In strategy 3, the similarity propagation is between two graphs combined by semantic subgraphs. 
To balance the above two strategies, we propose a hybrid propagation scale strategy. 
In this strategy, one graph is a semantic subgraph of an element $e_i$, and the another graph is the combined graph $G_{C}$. 
It means that the similarity propagation is between $G_{Si}^A$ and $G_C^B$ or between $G_C^A$ and $G_{Sj}^B$. 
Therefore, after one similarity propagation, we can obtain the similarities about $e_i$ to all elements in another ontology. 
This strategy only needs $n$ times propagations.

For hybrid semantic subgraph propagation scale strategy, given an element $a_i$, we can get a set of similarities \{$S(a_i,b_x)$\}$(x=1,...,n)$ in the similarity propagation. 
Given another element $b_j$ in the opponent ontology, we can also get another similarity set
$\{S(a_y,b_j)\}(y=1,...,m)$. 
Therefore, we have two similarity matrices, in which the similarity value at $S(a_i,b_j)$ may be different.
This paper calculates the average of two similarity matrices as the final propagation result.
The two similarity matrices have the function of cross validation, so it can improve the quality of propagation result.



\section{Matcher based on Semantic Subgraphs}
Moreover, we propose a matcher based on semantic subgraphs for inputting credible aliment seeds to similarity propagation. 
This matcher first constructs the semantic description document for each element. Then it calculates the similarities between elements based on the semantic description documents. 
\subsection {Semantic description document}

For an element, this paper organizes relevant literal information based on semantic subgraphs as virtual document \cite{Qu2006}. 
We call this virtual document the \emph{semantic description document} (SDD). 
To avoid introducing irrelevant literal information, SDD is constrained in semantic subgraphs. 
In addition, SDD does not consider ontology language primitive, such as \emph{rdfs:Class} and \emph{owl:hasValue}. 
In SDD construction, the text preprocessing contains stemming and removing frequent vocabularies.

For each concept, property or instance, it has a basic SDD, which consists of local name, label and annotation. 
The basic SDD of element $e$ is: 
\begin{eqnarray}
{D_{base}}(e) &=& {\varphi _1}*{W_{localname}} + {\varphi _2}*{W_{label}} + {\varphi _3}*{W_{comment}}   
{}+ {\varphi _4}*{W_{otherAnnotation}}
\end{eqnarray}
where $W_{localname}$ is local name, $W_{label}$ is \emph{rdfs:label} text, $W_{comment}$ is the \emph{rdfs:comment} text, and $W_{otherAnnotation}$ is other annotation text. 
Weight $\varphi _i$  is in [0,1]. 
Hence, SDD is the set of words with weights. 
$+$ denotes the union operation between sets.

Since an ontology graph is enriched, elements with \emph{owl:equivalentClass} or \emph{owl:sameAs} axioms will have same SDD.

Generally, we consider the SDD in two sides: 
(1) SDD can re-organize literal information according to the semantic description of elements; 
(2) To avoid containing irrelevant and unimportant literal information, SDD is constrained in semantic subgraphs.  


The SDD of concept $C$ is organized by concept hierarchy, axioms, related properties and instances.
Three virtual documents are constructed to describe the semantic context for sup-concepts, sub-concepts and sibling concepts. 
$sup(C)$, $sub(C)$ and $sib(C)$ are sets of super-concepts, sub-concepts and sibling concepts of $C$, respectively. 
$dist(C, C_i)$ is the distance between $C$ and $C_i$.
\begin{eqnarray}
D{_{\sup }}(C) &= &\sum\limits_{{C_i} \in \sup (C)} {{1 \over {dist(C,{C_i})}}} D{_{base}}({C_i}) \\
D{_{sub}}(C) &= &\sum\limits_{{C_i} \in sub(C)} {{1 \over {dist(C_i)}}} D{_{base}}({C_i}) \\
D_{sib}(C)&=&\sum\limits_{{C_i} \in sib(C)}{D_{base}(C_i)}
\end{eqnarray}

Virtual documents of related properties of concept $C$ are follows.
\begin{eqnarray}
D_{dom}(C) = \sum\limits_{{P_i} \in Dom^{*}(C)} D_{base}(P_i) \\
D_{rng}(C) = \sum\limits_{{P_i} \in Rng^{*}(C)} D_{base}(P_i)
\end{eqnarray}
where $Dom^{*}(C)$ and $Rng^{*}(C)$ are properties whose domain and range are $C$.

A virtual document is used to describe direct instances of $C$:
\begin{equation}
D_{ins}(C) =  \sum\limits_{{c_i} \in C} D_{base}(c_i)
\end{equation}

The SDD of property $P$ is organized by domain and range statements, and it contains two parts:
\begin{eqnarray}
D{_{dom}}(P) = \sum\limits_{{C_i} \in Dom(P)} {D{_{base}}({C_i})} \\
D{_{rng}}(P) = \sum\limits_{{e_i} \in Rng(P)} {D{_{base}}({e_i})} 
\end{eqnarray}


An ontology often has blank nodes, which are contained by SDD of concepts and properties. 
The SDD of a blank node $b$ is as follows.
\begin{eqnarray}
  D{_{blank}}(b)=
  && {\alpha _1}*\sum\limits_{{t_i}\in C1} {D{_{base}}(pre({t_i})) + D{_{base}}(obj({t_i}))}   \nonumber\\
  && + {\alpha _2}*\sum\limits_{{t_l}\in C2} {D{_{base}}(sub({t_l})) + D{_{base}}(pre({t_l}))}   \nonumber\\
  && + {\alpha _3}*\sum\limits_{{t_m}\in C3} {D{_{base}}(pre({t_m})) + D{_{blank}}(obj({t_m}))}   \nonumber\\
  && + {\alpha _4}*\sum\limits_{{t_n}\in C4} {D{_{blank}}(sub({t_n})) + D{_{base}}(pre({t_n}))} 
\end{eqnarray}
Let $B$ be the set of blank nodes in ontology $O$.
Given a triple  $t$ =$<s,p,o>$, $sub(t)$, $pre(t)$ and $obj(t)$ are  subject, predicate and object. 
 $C1$, $C2$, $C3$ and $C4$ are sets of triples: 
\begin{eqnarray*}
C1&=&\{t \in O| sub(t)=b \ and\  obj(t) \notin B\}\\
C2&=&\{t \in  O| sub(t) \notin B \ and\ obj(t)=b\}\\
C3&=&\{t \in O| sub(t)=b \ and\ obj(t) \in B\}\\
C4&=&\{t \in  O| sub(t) \in B \ and\ obj(t)=b\}
\end{eqnarray*}
$\alpha_i$ is a weight. 
Let $dist(t, b)$ be the distance between $t$ and $b$, then $\alpha_i=1/dist(t, b)$.
$D{_{blank}}(b)$ can be computed recursively. If there is a circle during recursive computing, the computing is terminated directly.

\subsection{Similarity computation}

After constructing SDD for concepts and properties, correspondences can be discovered by computing similarities between SDD. 
A SDD is a set of vocabularies with weights, namely, $SDD=\{p_1*W_1, p_2*W_2,...,p_x*W_x\}$. 
We can use cosine to measure the similarities.

Let $Doc=\{SDD_1,SDD_2,…,SDD_N\}$, and each $SDD$ contains $n$ items $t_1,t_2,…,t_n$. 
Thus each document $SDD_i$ can be described as an $n$-dimension vector $\vec D_i=(d_{i1},d_{i2},...,d_{in})$, where $d_{ij}$ is the weight of $j$-th item. 
If the edit-distance similarity of two items is larger than a predefined threshold 0.85, they are treated as same item. 
The weight $d_{ij}$ in vector $\vec D_i$ is TF-IDF weight. 

The similarity between two virtual documents is the cosine value of vectors. Therefore, the similarity between $\vec D_i$ and $\vec D_j$ is: 
\begin{eqnarray}
Sim({\vec D_i},{\vec D_j}) =  {{\sum\limits_{k = 1}^n {{d_{ik}} \times {d_{jk}}} } \over {\sqrt {\sum\limits_{k = 1}^n {d_{ik}^2 \times \sum\limits_{k = 1}^n {d_{jk}^2} } } }}
\end{eqnarray}

In addition, for the reason that we divide and organize all literal information according to semantics, this matcher based on semantic subgraphs also performs well for informative ontologies.

\section{Experimental Evaluation}

We have implemented the method for matching weak informative ontologies in our ontology matching system Lily \footnote{https://github.com/npubird/LilyWIO}. 
Lily is implemented in Java and C++. 
In this section, we first present the dataset, criteria, and settings used in the experiments. Secondly, we verify the proposed method and compare with other works. Thirdly,  we discuss the effect of propagation scale strategies. Then we address the influence of initial similarity seeds and propagation performance.  Finally, we discuss the results on general ontology matching tasks.

\subsection{Dataset, criteria and  settings}

In the evaluation, we use  OAEI benchmark as the dataset.  
The reason is that the OAEI benchmark includes not only informative ontologies but also some WIOs, which are very similar to the WIOs used in practical applications.
From 2004 to 2016, even though OAEI benchmark has some changes in each year, it has little effect on the fairness of evaluation.
This paper uses benchmark2008 and benchmark2009\footnote{http://oaei.ontologymatching.org/2009/benchmarks/}, which are two similar versions of the OAEI benchmark. 
The dataset has 110 matching tasks  including 50 basic matching tasks and 60 transformations from basic tasks. 
All matching tasks have non-sequential number from 101 to 304.

According to characteristics of the dataset, we divide it into five groups: 
\begin{itemize}
\item 101-104: This group contains same, irrelevant, language generalized and restricted ontologies. 
\item  201-210: In this group, ontology structure is preserved, but labels and identifiers are replaced by random names, misspellings, synonyms and foreign names. 
The comments have been suppressed in some cases.
These ontologies are similar to WIOs in industrial applications. 
\item 221-247: This group is divided into two subgroups: 221-231 and 232-247. The first subgroup contains 11 kinds of modifications. For example, the hierarchy is flattened or expanded, and individuals, restrictions and data types are suppressed. 
In the second subgroup, the modifications are the combinations of the ones used in 221-231.
\item 248-266: Most ontologies in this group are weak informative ontologies. 
All labels and identifiers are replaced by random names, and  comments are also suppressed. 
Therefore, all ontologies in this group are WIOs.
\item 301-304: This group contains 4 real-world matching tasks.
\end{itemize}

We inspect the dataset manually and select the 78 WIOs as shown  in Table~\ref{ov_dataset}. 
There is no WIOs in three groups: 101-103, 221-247 and 301-304. 
All 67 ontologies in group 248-266 are WIOs. 
There are 11 WIOs in group 201-210. 
Therefore, the ratio of WIOs in this datasets is $78/110=70.9\%$.
These WIOs will be used to verify our method.

\begin{table}
	\caption{Overview of OAEI benchmark(2008,2009) dataset}
	\label{ov_dataset}
	\begin{center}
	\begin{tabular} {lll}
			\hline
			\noalign{\smallskip} 
			group& number of ontologies & weak informative ontologies\\
			\noalign{\smallskip}
			\hline

			101-104	&3		& N/A\\
			201-210	&18		& 201-6,201-8,202,202-4,202-6,202-8,205,206,207,209,210 \\
			221-247	&18		& N/A\\
			248-266	&67		&all ontologies \\
			301-304	&4		& N/A\\

		\hline
		\end{tabular}
	\end{center}
\end{table}

This paper uses the classical criteria: precision, recall and F1-measure to evaluate the matching results. 
Let $Q$ be the alignment of our method and $T$ be the reference alignment,  then the precision, recall and F1-Measure are:
\begin {eqnarray}
P=\frac{|Q\cap  T|}{|Q|} \\ 
R=\frac{|Q\cap T|}{|T|} \\
\text{F1-measure}=\frac{2PR}{P+R}
\end{eqnarray}

In the evaluation, we set $ \lambda=0.85$ in circuit model, $\theta\!=\!0.005$ in SC-condition, and  $\phi=\delta=0.25$ for checking weak informative ontologies.

\subsection{Over performance on weak informative ontologies}

We verify our method on the 78 weak informative ontology matching tasks listed in Table~\ref{ov_dataset}. 
In order to simplify results, we do not list the matching result for each task, but use the prefix numbers to divide the 78 matching tasks into 22 groups. 
For each group, we present the average matching results. 
For example, the 202 group contains 4 matching tasks with prefix 202, namely, 202, 202-4, 202-6 and 202-8, and we calculate the average precision, recall and F1-measure on the 4 matching tasks. 
In addition, in the similarity propagation model,  we use the hybrid semantic subgraph propagation strategy.

\begin{figure*}[htb!]

	\begin{centering}
		\includegraphics[trim = 0mm 0mm 0mm 0mm, clip,  width=1.00\textwidth]{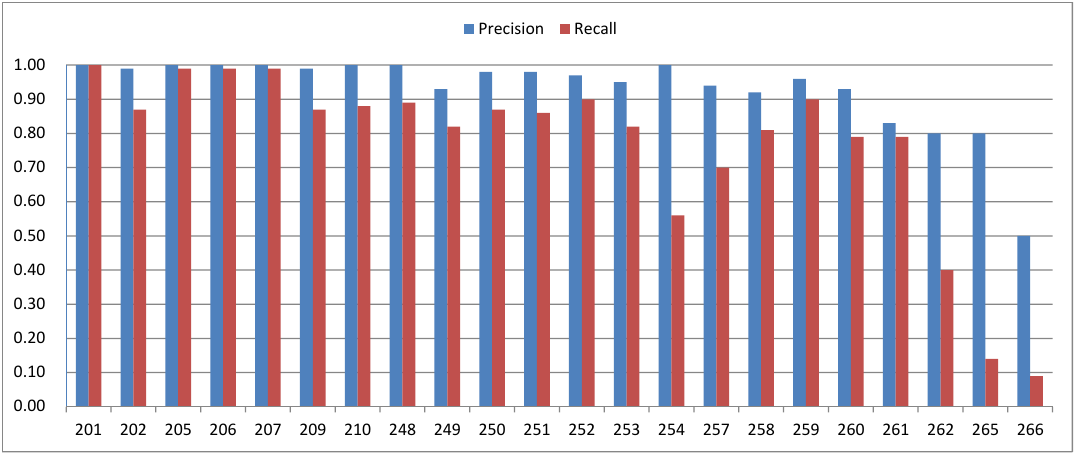}
		\caption{Matching results on weak informative ontologies}
			\label{mr_wio}
	\end{centering}
\end{figure*}

Figure~\ref{mr_wio} presents the matching results obtained by our method on the 22 groups. 
For group 201-210 , our method produces high quality results, whose precisions are larger than 0.99 and recalls are larger than 0.87. 
The results mean that if the ontology structure is preserved, misspellings, synonyms, foreign names, and even labels and identifiers are replaced by random names, our method still works perfectly. 
The explanation for this fact is that our method can utilize limited literal information to generate few credible seeds and then find more correct alignments by the similarity propagation model. 
Although all matching tasks in group 248-266 are difficult, our method also perform well on them. 
For most tasks in group 248-266, the precisions are larger than 0.8 and the recalls are larger than 0.7. 
For group 254, 262, 265 and 266, we cannot obtain results with high recalls. 
The main reason is that the ontology structure is not preserved, especially, the concept hierarchy is flattened and there is no property. 
Moreover, comments are removed and labels are scrambled by random names. 
Therefore, the similarity propagation cannot work without structure information,  and there is, theoretically, no matching method can deal with this situation.

\begin{figure*}[htb!]

	\begin{centering}
		\includegraphics[trim = 0mm 0mm 0mm 0mm, clip, width=0.90\textwidth]{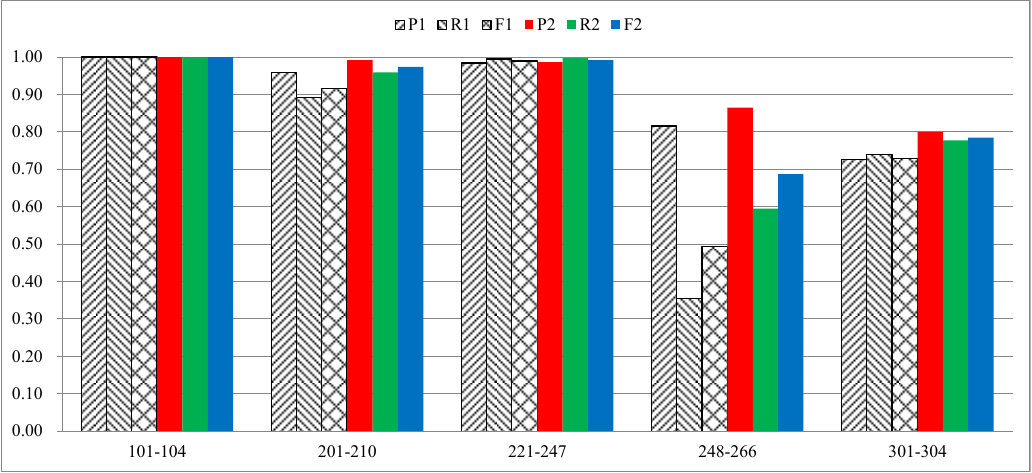}
		\caption{Matching results without similarity propagation VS with similarity propagation 
			(P1, R1 and F1 are the precision, recall and F1-measure for matcher based on semantic subgraphs;
			P2, R2 and F2 are the precision, recall and F1-measure for similarity propagation model)}
			\label{mr_spvs}
	\end{centering}
\end{figure*}

As shown in Figure~\ref{mr_spvs}, we compare the results with similarity propagation and the results without similarity propagation. 
The dataset used here is benchmark2008, which includes both informative ontologies and weak informative ontologies. 
We run our matcher based on semantic subgraphs (in Section 9) and the similarity propagation method (in Section 7 and 8), respectively. 
According to Figure~\ref{mr_spvs}, we observe the following facts.
\begin{itemize}
	\item The similarity propagation method proposed in this paper improves the quality of matching results, especially for the weak informative ontologies, such as the ontologies in group 248-266.
	\item For the weak informative ontologies, our similarity propagation method can increase the recall of results greatly. 
	The reason is that our similarity propagation model can discover more correct correspondences with few correspondences as seeds. 
	Especially, these new correspondences are difficult to be discovered by the methods without similarity propagation.
	Therefore, our similarity propagation model can improve the recall of matching results.
	\item For informative ontologies, such as group 221-247, our method can also produces good results. It means that our method is a general matching method.
\end{itemize}

Table~\ref{c_b2018} and Table~\ref{c_b2019} compare the results obtained by different matching systems on the 22 group of weak informative ontologies in benchmark2008 and benchmark2009, respectively.  
Lily is our matching system with the method for matching WIOs. 
The first column lists the 22 group matching tasks. 
From the second column, each column is the F1-measure results of one system. 
The last row is the average F1-measure of each system.
Especially, we compare similarity flooding (SF) algorithm with other ontology matching systems. 

Table~\ref{c_b2018} shows the results of 17 systems, in which the results of the first 13 systems are retrieved from http://oaei.ontologymatching.org/2008/results, 
and results of AML\footnote{https://github.com/AgreementMakerLight/AML-Project} and LogMap \footnote{https://github.com/ernestojimenezruiz/logmap-matcher} are obtained by running original source codes, respectively. 
SF stands for the results of the similarity flooding algorithm.
Not all systems perform well on the weak informative ontologies matching tasks. 
In fact, 11 in 17 systems are smaller than 0.6 F1-measure, and 5 systems are in 0.6 to 0.8 F1-measure. 
Only Lily obtains the highest 0.82 F1-measure. 
Specially, Lily obtains the best matching results for most matching tasks.

Besides Lily, some systems in Table 4 (such as ASMOV, aflood, RIMOM, SAMBO, GeRoMe) have their similarity propagation methods. 
They also have good performance on the weak informative ontologies. 
It demonstrates that similarity propagation is an efficient way to match weak informative ontologies.

Table~\ref{c_b2019} shows the matching results of 16 systems on weak informative matching tasks, and the first 15 results are obtained from  http://oaei.ontologymatching.org/2009/results.  Results of AML and LogMap is obtained by running source codes. 
It also can be seen that Lily is the best system. 
Specifically, the F1-measure of 9 systems is small than 0.6, and 6 systems have the F1-measure in 0.6 to 0.8. 
Only Lily and ASMOV get 0.82 and 0.8 F1-measure, respectively. 
For most matching tasks, Lily also obtains the best F1-measure.

\begin{table}[htb!]

	\caption{Comparison of 17 systems on weak informative ontologies in benchmark2008 (F1-measure)}
		\label{c_b2018}
	\begin{center}
	\resizebox{\textwidth}{!}{	
	\begin{tabular}{ccccccccccccccccccc} 
			\hline
			\noalign{\smallskip} 
			&{edna}	 &{aflood} &{aroma}	&{ASMOV} &{CIDER} &{DSSim}	&{GeRoMe} &{TaxoMap} &{MapPSO} &{RiMOM} &{SAMBO} &{SAMBOdtf} &{SPIDER} &AML &LogMap &SF &{\textbf{Lily}} \\
			\noalign{\smallskip}
			\hline
201	&0.32 	&0.83 	&0.99 		&\textbf{1.00} 		&0.85 	&0.95 		&0.86 	&0.27 	&0.41 	&\textbf{1.00} 	&0.65 	&0.65 	&0.85 			&0.52 &0.48 &0.32	& \textbf{1.00}\\ 
202	&0.32 	&0.84 	&0.74 		&0.90 		&0.46 	&0.62 		&0.55 	&0.15 	&0.30 	&\textbf{0.93} 	&0.43 	&0.43 	&0.46 					  &0.58 &0.00 &0.25	& \textbf{0.93}\\
205	&0.34 	&0.69 	&\textbf{0.99} 	&\textbf{0.99} 	&0.85 	&0.84 		&0.88 	&0.16 	&0.31 	&\textbf{0.99} 	&0.57 	&0.61 	 &0.85 		 &0.41 	&0.38 &0.20	& \textbf{0.99}\\
206	&0.51 	&0.90 	&\textbf{0.99} 	&\textbf{0.99} 	&0.76 	&0.92 		&0.85 	&0.00 	&0.36 	&\textbf{0.99} 	&0.61 	&0.71 	&0.76 		 &0.12 &0.40 &0.16	& \textbf{0.99}\\
207	&0.51 	&0.90 	&\textbf{0.99} 	&\textbf{0.99} 	&0.74 	&0.93 		&0.85 	&0.00 	&0.37 	&\textbf{0.99} 	&0.61 	&0.71 	&0.74 		 &0.12 &0.40 &0.16	& \textbf{0.99}\\
209	&0.33 	&0.64 	&0.71 		&0.89 		&0.41 	&0.54 		&0.51 	&0.06 	&0.23 	&0.86 	&0.40 	&0.47 	&0.41 								&0.36	&0.00 &0.00 & \textbf{0.92}\\
210	&0.52 	&0.87 	&0.79 		&0.93 		&0.50 	&0.56 		&0.66 	&0.08 	&0.19 	&0.92 	&0.37 	&0.59 	&0.50 								&0.36	&0.00 &0.00 &\textbf{0.94} \\
248	&0.41 	&0.59 	&0.72 		&0.89 		&0.52 	&0.67 		&0.47 	&0.19 	&0.37 	&\textbf{0.93} 	&0.52 	&0.52 	&0.52 					 &0.49 &0.00 &0.22	& \textbf{0.93}\\
249	&0.40 	&0.76 	&0.54 		&0.87 		&0.55 	&0.51 		&0.54 	&0.19 	&0.38 	&0.86 	&0.52 	&0.52 	&0.55 								&0.50	&0.47 &0.50 & \textbf{0.89}\\
250	&0.23 	&0.74 	&0.79 		&0.87 		&0.60 	&0.70 		&0.66 	&0.42 	&0.42 	&0.91 	&0.52 	&0.52 	&0.60 								&0.00	&0.48 &0.00 & \textbf{0.93}\\
251	&0.40 	&0.73 	&0.76 		&0.88 		&0.54 	&0.68 		&0.55 	&0.17 	&0.37 	&0.83 	&0.52 	&0.52 	&0.54 								&0.50	&0.00 &0.00 & \textbf{0.91}\\
252	&0.62 	&0.73 	&0.82 		&0.93 		&0.71 	&0.79 		&0.59 	&0.28 	&0.54 	&0.92 	&0.71 	&0.71 	&0.71 								&0.69	&0.00 &0.69 & \textbf{0.94}\\
253	&0.41 	&0.60 	&0.49 		&0.84 		&0.52 	&0.51 		&0.47 	&0.19 	&0.38 	&0.82 	&0.52 	&0.52 	&0.52 								&0.49	&0.46 &0.49 & \textbf{0.87}\\
254	&0.23 	&\textbf{0.70} 	&0.63 	&\textbf{0.70} 	&0.52 	&\textbf{0.70} 	&0.09 	&0.43 	&0.40 	&\textbf{0.70} 	&0.52 	&0.52 	&0.52 	&0.52 &0.46 &0.37	& \textbf{0.70}\\
257	&0.23 	&0.55 	&0.60 		&0.68 		&0.60 	&0.51 		&0.66 	&0.42 	&0.41 	&\textbf{0.75} 	&0.52 	&0.52 	&0.60 					 &0.54 &0.48 &0.38	& 0.73\\
258	&0.40 	&0.73 	&0.51 		&0.80 		&0.54 	&0.52 		&0.54 	&0.17 	&0.38 	&0.68 	&0.52 	&0.52 	&0.54 								&0.50 &0.45 &0.25	& \textbf{0.86}\\
259	&0.63 	&0.71 	&0.69 		&0.89 		&0.71 	&0.72 		&0.59 	&0.28 	&0.56 	&0.85 	&0.70 	&0.70 	&0.71 								&0.69 &0.68 &0.38	& \textbf{0.93}\\
260	&0.20 	&0.75 	&0.73 		&0.81 		&0.56 	&0.71 		&0.50 	&0.42 	&0.40 	&\textbf{0.85} 	&0.51 	&0.51 	&0.56 					&0.52 &0.46 &0.41	& \textbf{0.85}\\
261	&0.32 	&0.80 	&0.77 		&0.81 		&0.71 	&0.76 		&0.61 	&0.58 	&0.45 	&0.76 	&0.68 	&0.68 	&0.71 								&0.72 &0.65 &0.40	& \textbf{0.83}\\
262	&0.23 	&\textbf{0.52} 	&0.45 	&0.51 		&\textbf{0.52} 	&0.51 	&0.09 	&0.43 	&0.41 	&\textbf{0.52} 	&\textbf{0.52} 	&\textbf{0.52} 	&\textbf{0.52} 	&\textbf{0.52} &0.46 &0.35	& \textbf{0.52}\\
265	&0.03 	&0.12 	&0.00 		&0.12 		&0.00 	&0.00 		&0.19 	&0.00 	&0.03 	&0.18 	&0.00 	&0.00 	&0.00 							 &0.00 &0.00 &0.12	& \textbf{0.24}\\
266	&0.02 	&0.05 	&0.00 		&0.10 		&0.00 	&0.00 		&0.08 	&0.00 	&0.02 	&0.00 	&0.00 	&0.00 	&0.00 							 &0.00 &0.00 &0.00	 & \textbf{0.14}\\
\hline
Avg.	&0.35 	&0.67 	&0.67 		&0.79 		&0.55 	&0.62 		&0.54 	&0.22 	&0.35 	&0.78 	&0.50 	&0.52 	&0.55 						&0.42 &0.31 &0.26	& \textbf{0.82}\\

		\hline
	\end{tabular}
 }
	\end{center}
\end{table}

\begin{table*}[htb!]

	\caption{Comparison of 16 systems on weak informative ontologies in benchmark2009 (F1-measure)}
		\label{c_b2019}
	\begin{center}
	\resizebox{\textwidth}{!}{		
	\begin{tabular} {ccccccccccccccccc}
			\hline
			\noalign{\smallskip} 
			&{edna}	 &{aflood} &{AgrMaker} &{aroma} &{ASMOV} &{DSSim} &{GeRoMe}	&{kosimap} &{MapPSO} &{RiMOM} &{SOBOM} &{TaxoMap} &AML &LogMap &SF &{\textbf{Lily}}\\	
			\noalign{\smallskip}
			\hline
201	&0.32 	&0.94 	&0.94 	&0.99 	&\textbf{1.00} 	&0.89 	&0.97 	&0.67 	&\textbf{1.00} 	&\textbf{1.00} 	&0.53 	&0.26 				&0.52 &0.48 &0.34 &\textbf{1.00} \\
202	&0.31 	&0.92 	&0.43 	&0.72 	&0.90 	&0.62 	&0.67 	&0.45 	&0.72 	&\textbf{0.94} 	&0.25 	&0.19 									&0.42 &0.00 &0.25 &0.92 \\
205	&0.34 	&0.82 	&0.98 	&0.99 	&\textbf{0.99} 	&0.86 	&0.98 	&0.69 	&\textbf{0.99} 	&\textbf{0.99} 	&0.45 	&0.15 				&0.41 &0.38 &0.25 &\textbf{0.99} \\
206	&0.52 	&0.95 	&0.93 	&\textbf{0.99} 	&\textbf{0.99} 	&0.77 	&0.95 	&0.77 	&\textbf{0.99} 	&\textbf{0.99} 	&0.00 	&0.10 	&0.00 &0.40 &0.16 &\textbf{0.99} \\
207	&0.52 	&0.95 	&0.93 	&\textbf{0.99} 	&\textbf{0.99} 	&0.77 	&0.95 	&0.77 	&\textbf{0.99} 	&\textbf{0.99} 	&0.00 	&0.10 	&0.12 &0.40 &0.20	&\textbf{0.99} \\
209	&0.35 	&0.80 	&0.35 	&0.73 	&0.90 	&0.53 	&0.70 	&0.48 	&0.67 	&0.88 	&0.17 	&0.09 											 &0.36 &0.00 &0.16	&\textbf{0.93} \\
210	&0.52 	&0.94 	&0.33 	&0.77 	&\textbf{0.97} 	&0.53 	&0.73 	&0.70 	&0.69 	&0.87 	&0.00 	&0.00 									&0.36 &0.00 &0.18 &0.94 \\
248	&0.41 	&0.82 	&0.52 	&0.71 	&0.90 	&0.68 	&0.71 	&0.52 	&0.36 	&0.87 	&0.32 	&0.22 											&0.32 &0.00 &0.24	&\textbf{0.94} \\
249	&0.41 	&\textbf{0.91} 	&0.52 	&0.51 	&0.84 	&0.52 	&0.59 	&0.53 	&0.07 	&0.84 	&0.33 	&0.24 								&0.50 &0.47 &0.26	&0.90 \\
250	&0.24 	&\textbf{1.00} 	&0.52 	&0.77 	&0.88 	&0.70 	&0.77 	&0.54 	&0.64 	&0.91 	&0.52 	&0.51 								&0.07  &0.48 &0.40 	&0.91 \\
251	&0.40 	&0.78 	&0.52 	&0.74 	&0.92 	&0.69 	&0.71 	&0.50 	&0.66 	&0.87 	&0.31 	&0.22 										&0.50  &0.00 &0.28	&\textbf{0.91} \\
252	&0.62 	&0.84 	&0.71 	&0.80 	&\textbf{0.94} 	&0.80 	&0.84 	&0.69 	&0.72 	&0.86 	&0.49 	&0.34 								&0.69 &0.00 &0.30 &\textbf{0.94} \\
253	&0.40 	&0.69 	&0.52 	&0.48 	&0.82 	&0.52 	&0.58 	&0.52 	&0.08 	&0.75 	&0.32 	&0.22 										&0.50 &0.46 &0.22	&\textbf{0.88} \\
254	&0.22 	&\textbf{0.70} 	&0.52 	&0.60 	&\textbf{0.70} 	&\textbf{0.70} 	&0.66 	&0.49 	&0.36 	&\textbf{0.70} 	&0.52 	&0.48 	&0.52 &0.46 &0.37	&\textbf{0.70} \\
257	&0.23 	&\textbf{0.96} 	&0.52 	&0.54 	&0.68 	&0.51 	&0.61 	&0.54 	&0.64 	&0.72 	&0.52 	&0.51 								&0.54 &0.48 &0.38	&0.75 \\
258	&0.40 	&0.69 	&0.52 	&0.49 	&0.84 	&0.52 	&0.58 	&0.50 	&0.15 	&0.68 	&0.31 	&0.22 										&0.50  &0.45 &0.28	&\textbf{0.86} \\
259	&0.62 	&0.78 	&0.71 	&0.69 	&0.90 	&0.73 	&0.74 	&0.69 	&0.19 	&0.67 	&0.49 	&0.34 									&0.69 &0.68 &0.32	&\textbf{0.92} \\
260	&0.19 	&0.78 	&0.51 	&0.68 	&0.81 	&0.71 	&0.73 	&0.53 	&0.45 	&0.79 	&0.52 	&0.51 									&0.52  &0.46 &0.45	&\textbf{0.85} \\
261	&0.31 	&0.84 	&0.69 	&0.71 	&\textbf{0.80} 	&0.76 	&0.78 	&0.70 	&0.45 	&0.75 	&0.71 	&0.69 							&0.72 &0.65 &0.42	&\textbf{0.80} \\
262	&0.23 	&\textbf{0.52} 	&\textbf{0.52} 	&0.44 	&\textbf{0.52} 	&0.51 	&0.51 	&0.49 	&0.42 	&0.51 	&\textbf{0.52} 	&0.48 	&0.52 &0.46 &0.34	&0.51 \\
265	&0.02 	&\textbf{0.24} 	&0.00 	&0.00 	&0.12 	&0.00 	&0.00 	&0.13 	&0.10 	&0.15 	&0.00 	&0.00 	&0.00 &0.00 &0.00	&\textbf{0.24} \\
266	&0.02 	&0.11 	&0.00 	&0.00 	&0.10 	&0.00 	&0.00 	&0.11 	&0.06 	&0.09 	&0.00 	&0.00 	&0.00  &0.00 &0.00	&\textbf{0.15} \\
		\hline
Avg.	&0.35 	&0.77 	&0.55 	&0.65 	&0.80 	&0.61 	&0.67 	&0.54 	&0.52 	&0.76 	&0.33 	&0.27 							&0.40 &0.31 &0.26	&\textbf{0.82} \\
		\hline
	\end{tabular}
    }
	\end{center}

\end{table*}

It should be noted that the similarity flooding has very poor performance on the datasets. 
The results support our analysis that similarity flooding cannot directly deal with ontology matching. 

Finally, we study some alignment cases in Table~\ref{c_b2018}. 
First, we randomly select 4 alignments  obtained by Lily in 210 task, which is a cross-lingual ontology matching task. 
The selected alignments are:
\emph{MastersThesis} = \emph{MémoireDeMastère},
\emph{PhdThesis} = \emph{MémoireDeDoctorat},
\emph{numberOrVolume} = \emph{numéroOuVolume},
\emph{LectureNotes} = \emph{Polycopié}.
It is surprising that Lily discovers these alignments without any external cross-lingual lexicon or resource. 
Lily only uses similarity propagation with few seeds to find these correct alignments, and it performs better than some systems with string-based matchers.
Then, we randomly select 4 alignments obtained by Lily in 248 task, in which most literal information is replaced by meaningless strings. 
The selected alignments are:
\emph{numberOrVolume} = \emph{dzezd},
\emph{LectureNotes} = \emph{scds},
\emph{proceedings} = \emph{zassdzadb},
\emph{institution} = \emph{hsgiuyza}.
These results are also existing. 
Here, the initial seeds contain few alignments such as \emph{Address} = \emph{Address} and \emph{lastName} = \emph{lastName}, then more and more correct alignments are got by similarity propagation.
The examination demonstrates that similarity propagation is a very effective way to deal with weak informative ontologies.  
In the inverse perspective, we find some alignments obtained by string-based systems, but the alignments are covered by results of Lily.
The reason is that the matcher based on semantic subgraphs in Lily is also an effective string-based matcher. 
Lily uses this matcher to provide initial seeds in similarity propagation.

Therefore, the above experimental results demonstrate that our method for matching weak informative ontologies is very effective, and the matching system with our method can perform better than other systems.

\subsection{Evaluating propagation scale strategies }

This experiment aims to evaluate the performance of different propagation scale strategies.
The results are obtained on the 248 matching task, which is selected randomly. 
Similar experimental results and conclusions can also be obtained on any other weak informative matching tasks. 
In this experiment, we use S1, S2, S3, S4 and S5 to represent five kinds of propagation scale strategies mentioned in Section 8. 

In order to demonstrate the influence of sizes of semantic subgraphs, we change sizes of semantic subgraphs from 0 to 35. 
A 0-size semantic subgraph only contains an element, that means the matching method only uses the information of the element. 
Introducing the change of the semantic subgraph size will help us to know how to set suitable size for semantic subgraphs in the similarity propagation.

Table~\ref{c_dps} shows the experimental results (F1-measure) for comparing different strategies.  
$Size$ column is the semantic subgraph size. 
$Seed$ denotes the F1-measure of initial similarity seeds obtained by the matcher based on semantic subgraphs (in Section 9).
Other columns are the results obtained by different propagation scale strategies.
The last row provides the average F1-measure for semantic subgraphs from size 0 to 35. 

\begin {table*}
\centering
\caption {Comparison of different propagation scale strategies (F1-measure)}
\label{c_dps}
\begin{tabular}{p{1.1cm}p{1.1cm}p{1.1cm}p{1.1cm}p{1.1cm}p{1.1cm}p{1.1cm}}
	\hline
	\noalign{\smallskip} 
	Size & Seed & S1 & S2 & S3 & S4 & S5   \\
	\noalign{\smallskip}
	\hline
	0 & 0.02 & 0.02 & 0.02 & 0.02 & 0.02 & 0.02   \\
	1 & 0.37 & 0.60 & 0.60 & 0.42 & 0.25 & 0.50   \\
	2 & 0.43 & 0.65 & 0.60 & 0.55 & 0.35 & 0.55   \\
	3 & 0.42 & 0.53 & 0.48 & 0.54 & 0.40 & 0.74   \\
	5 & 0.49 & 0.61 & 0.53 & 0.66 & 0.61 & 0.80  \\
	10 & 0.54 & 0.59 & 0.59 & 0.66 & 0.69 & 0.83   \\
	15 & 0.59 & 0.64 & 0.59 & 0.66 & 0.66 & 0.84   \\
	20 & 0.56 & 0.61 & 0.63 & 0.67 & 0.68 & 0.79   \\
	25 & 0.56 & 0.63 & 0.60 & 0.66 & 0.73 & 0.83   \\
	30 & 0.56 & 0.65 & 0.69 & 0.66 & 0.71 & 0.88   \\
	35 & 0.56 & 0.62 & 0.65 & 0.62 & 0.65 & 0.83   \\
	\hline
	Avg. & 0.53 & 0.62 & 0.61 & 0.66 & 0.68 & 0.83   \\
	
	\hline
\end{tabular}

\end {table*}

According to the F1-measure of seeds, Table~\ref{c_dps} shows that the similarity propagation can improve the matching results.
The performances of similarity propagation scale strategies can be ranked as: 
$S5\!>\!S4\!>\!S3\!>\!S1\!>\!S2$. 
Further analyzing the experimental results, some interesting facts are observed as follows:
\begin{itemize}
	\item Even the full graph strategy improves the quality of matching results, it is not the best propagation scale strategy.
	\item It is surprising that the independent semantic subgraph strategy produces the worst matching results. 
	In this strategy, we find that some incorrect correspondences may have high similarity values during similarity propagation. 
	Since the independent semantic subgraphs are too small, the noise similarity values cannot be suppressed in propagation. 
	\item Combined semantic subgraph strategy (S3 and S4) can produce better results than above two strategies. 
	That may be explained by the fact that the two strategies not only remove some irrelevant information from ontologies, but also assure that some noise similarity values can be suppressed during propagation. 
	S3 and S4 have very close average F1-measure. 
	Therefore, two combined semantic subgraphs strategies can  improve the quality of results. 
	\item The hybrid semantic subgraph propagation strategy (S5) is the best strategy and produces the best results. 
	The reason is that this strategy combines advantages of S2, S3 and S4. 
	Moreover,  the cross validation function in S5 strategy also plays positive role.
	\item Finally, we also find that bigger semantic subgraphs cannot always produce better matching results. 
	Therefore, we set reasonable size for semantic subgraphs in the range 10 to 15.  
\end{itemize}

\subsection{Influence of initial similarity seeds}

The traditional similarity flooding model claims that initial similarity seeds would not affect propagation results. 
However, in our new model, we need to validate whether and how our similarity propagation method is sensitive to initial similarity seeds. 
In this experiment, we randomly modify seeds to satisfy the condition:
$Precision=Recall=F1$\emph{-measure}. 
The experimental data is also obtained on 248 matching task.
We let F1-measure of seeds increases from 0 to 1.0 with step 0.1, then observe changes of F1-measure with different propagation strategies. 

\begin{figure*}[htb!]

	\begin{centering}
		\includegraphics[trim = 0mm 0mm 0mm 0mm, clip, width=1.00\textwidth]{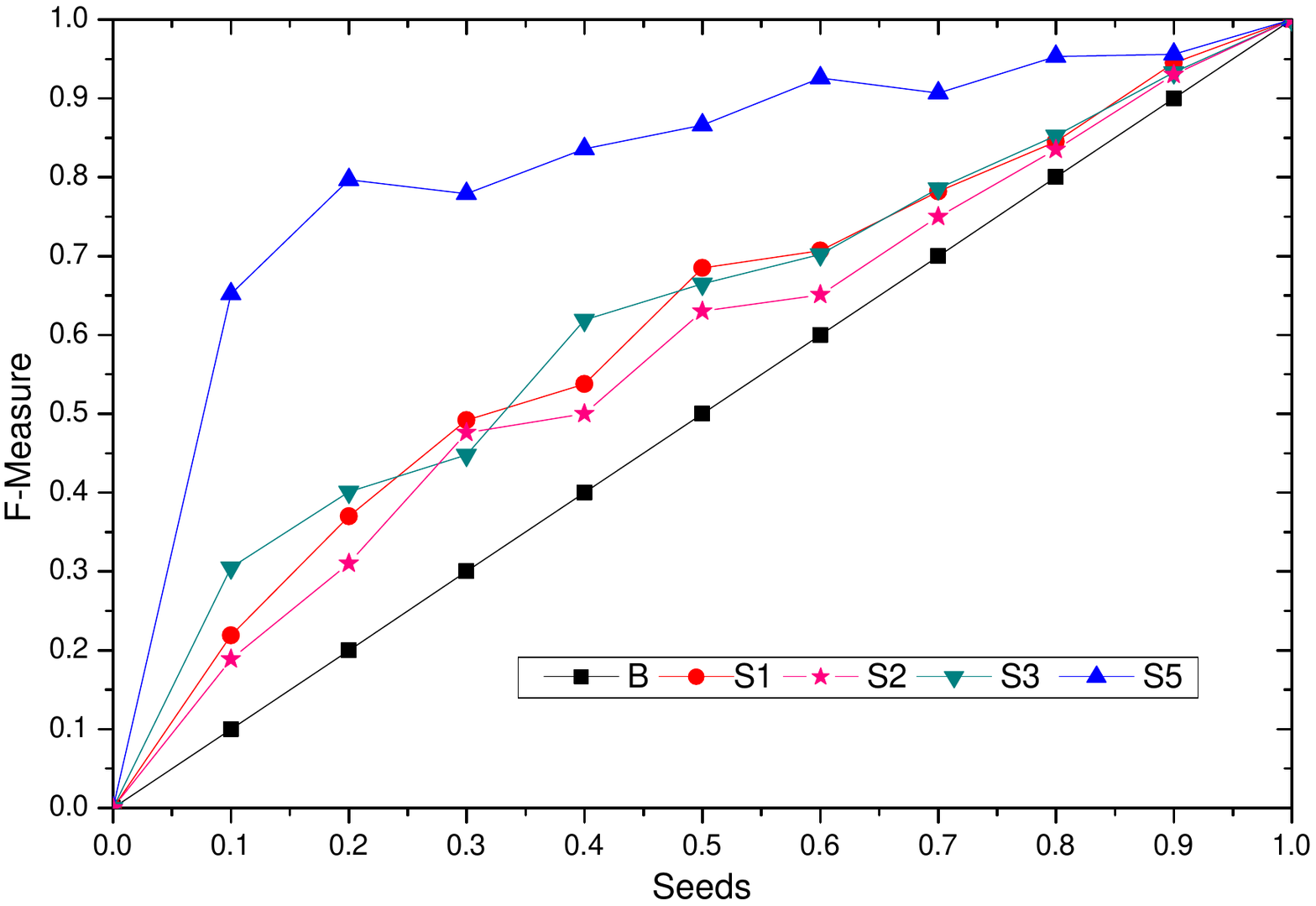}
		\caption{Influence of initial seed to matching results}
			\label{i_seed}
	\end{centering}
\end{figure*}

Figure~\ref{i_seed} shows the influence of initial similarity seeds to matching results. 
Line B denotes the F1-measure of seeds. 
S1, S2, S3 and S5 are F1-measure lines for corresponding propagation scale strategies. Since S4 has similar result to S3, we omit line S4. 

We observe some interesting facts in Figure~\ref{i_seed}:
\begin{itemize}
\item The initial seeds influence the matching result greatly. 
With the changes of seed F1-measure, matching results change monotonously. 
\item After similarity propagating, matching result is better than the initial seed. 
Meanwhile, better seeds would produce better matching results. 
\item Compared to other propagation scale strategies, S5 strategy is influenced by the seed slightly. 
It means that S5 strategy can use seeds with low F1-measure to produce higher quality results than other strategies. 
Therefore, S5 is the preferred propagation scale strategy.
\end{itemize}

\subsection{Propagation performance}

In the practical ontology matching tasks, we find that the similarity propagation would take up 30\%-50\% running time. 
The reason is that large semantic subgraphs would cause large pairwise connectivity graphs, and the iteration process for calculating fixpoint would also terminate slowly. 
Therefore, the semantic subgraph size is the key factor of performance.
We run our matching method with different propagation scale strategies on 248 task, then record the running time. Meanwhile, for each propagation scale strategy, the semantic subgraph size increases from 0 to 60 by 5 steps.  

\begin{figure*}[htb!]

	\begin{centering}
		\includegraphics[trim = 0mm 0mm 0mm 0mm, clip, width=0.90\textwidth]{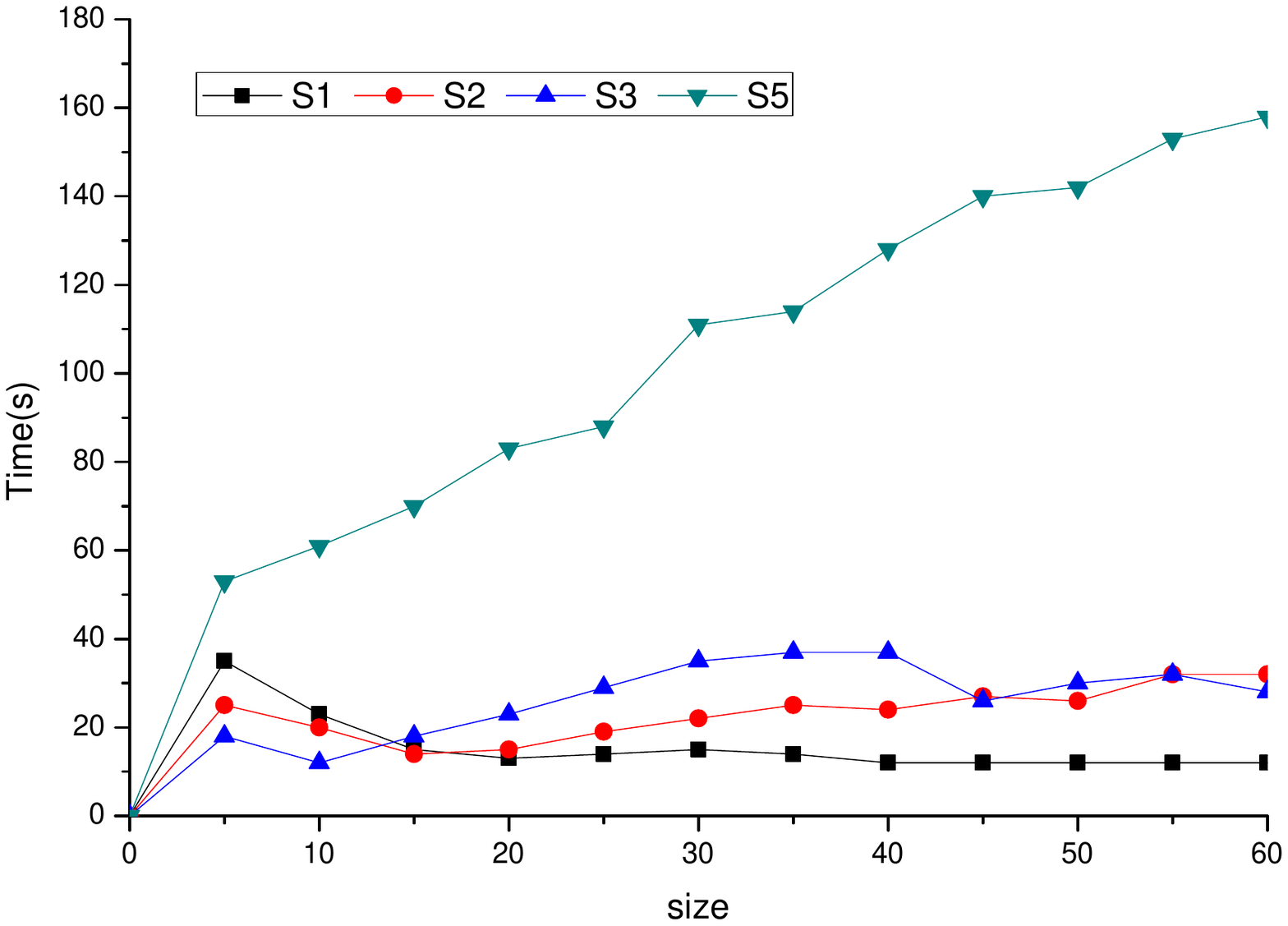}
		\caption{Performances of different propagation scale strategies}
			\label{p_ps}
	\end{centering}
\end{figure*}

Figure~\ref{p_ps} demonstrates the running time on various semantic subgraph sizes and different propagation scale strategies. We have some interesting facts as follows:
\begin{itemize}
\item For S1, S2 and S3 strategies, the running time has no relevant to the semantic subgraph sizes. 
When we change the semantic subgraph size, the lines of S1, S2 and S3 are almost steady. 
This result can be explained by the fact that the size of pairwise connectivity graph will keep stable when the semantic subgraph size is larger than 5. 
Therefore, we can infer that the running time of S1, S2 and S3 strategies have little correlation with the semantic subgraph size.
\item For the hybrid propagation scale strategy S5, Figure~\ref{p_ps} shows that the running time quickly increases with the semantic subgraph size increasing. 
There are two graphs in this strategy, one is the independent semantic subgraph for an element, and another is the combined semantic subgraph. 
When the semantic subgraph size increases, although the size of combined semantic subgraph keeps steady, the independent semantic subgraph will grow linearly. 
Therefore, the pairwise connectivity graph will grow too. 
Finally, it needs more time to run the similarity propagation.  
This fact also means that we should set suitable semantic subgraph size for hybrid propagation scale strategy.
\end{itemize}

\subsection{Performance  on general ontology matching tasks}

\begin{table*}
	\small
	\renewcommand\arraystretch{0.65}  
  \centering

  \caption{Matching results on benchmark2008}
    \label{gmr_b2008}
    \begin{tabular}{lllclllc}
    \hline 
     & \tiny{Precision} & \tiny{Recall}  & \tiny{F1-Measure} &  & \tiny{Precision} & \tiny{Recall}  & \tiny{F1-Measure} \\
    101   & 1.00  & 1.00  & 1.00  & 251*   & 0.96  & 0.76  & 0.85  \\
    103   & 1.00  & 1.00  & 1.00  & 251-2* & 0.99  & 0.96  & 0.97  \\
    104   & 1.00  & 1.00  & 1.00  & 251-4* & 0.99  & 0.90  & 0.94  \\
    \cline{1-4}
    201   & 1.00  & 1.00  & 1.00  & 251-6* & 0.94  & 0.83  & 0.88  \\
    201-2 & 1.00  & 1.00  & 1.00  & 251-8* & 0.99  & 0.85  & 0.91  \\
    201-4 & 1.00  & 1.00  & 1.00  & 252*   & 0.96  & 0.79  & 0.87  \\
    201-6* & 1.00  & 1.00  & 1.00  & 252-2* & 0.97  & 0.94  & 0.95  \\
    201-8* & 1.00  & 1.00  & 1.00  & 252-4* & 0.97  & 0.94  & 0.95  \\
    202*   & 1.00  & 0.84  & 0.91  & 252-6* & 0.97  & 0.94  & 0.95  \\
    202-2 & 1.00  & 0.97  & 0.98  & 252-8* & 0.97  & 0.94  & 0.95  \\
    202-4* & 1.00  & 0.92  & 0.96  & 253*   & 0.81  & 0.59  & 0.68  \\
    202-6* & 0.98  & 0.87  & 0.92  & 253-2* & 0.98  & 0.93  & 0.95  \\
    202-8* & 0.98  & 0.85  & 0.91  & 253-4* & 1.00  & 0.92  & 0.96  \\
    203   & 1.00  & 1.00  & 1.00  & 253-6* & 0.95  & 0.81  & 0.87  \\
    204   & 1.00  & 1.00  & 1.00  & 253-8* & 0.95  & 0.79  & 0.86  \\
    205*   & 1.00  & 0.99  & 0.99  & 254*   & 1.00  & 0.27  & 0.43  \\
    206*   & 1.00  & 0.99  & 0.99  & 254-2* & 1.00  & 0.82  & 0.90  \\
    207*   & 1.00  & 0.99  & 0.99  & 254-4* & 1.00  & 0.70  & 0.82  \\
    208   & 1.00  & 0.99  & 0.99  & 254-6* & 1.00  & 0.61  & 0.76  \\
    209*   & 0.97  & 0.88  & 0.92  & 254-8* & 1.00  & 0.42  & 0.59  \\
    210*   & 1.00  & 0.89  & 0.94  & 257*   & 0.50  & 0.06  & 0.11  \\
    \cline{1-4}
    221   & 1.00  & 1.00  & 1.00  & 257-2* & 1.00  & 0.97  & 0.98  \\
    222   & 1.00  & 1.00  & 1.00  & 257-4* & 0.94  & 0.88  & 0.91  \\
    223   & 0.98  & 0.98  & 0.98  & 257-6* & 0.84  & 0.79  & 0.81  \\
    224   & 1.00  & 1.00  & 1.00  & 257-8* & 0.89  & 0.76  & 0.82  \\
    225   & 1.00  & 1.00  & 1.00  & 258*   & 0.80  & 0.60  & 0.69  \\
    228   & 1.00  & 1.00  & 1.00  & 258-2* & 0.97  & 0.94  & 0.95  \\
    230   & 0.94  & 1.00  & 0.97  & 258-4* & 0.96  & 0.88  & 0.92  \\
    231   & 1.00  & 1.00  & 1.00  & 258-6* & 0.95  & 0.82  & 0.88  \\
    232   & 1.00  & 1.00  & 1.00  & 258-8* & 0.94  & 0.78  & 0.85  \\
    233   & 1.00  & 1.00  & 1.00  & 259*   & 0.89  & 0.70  & 0.78  \\
    236   & 1.00  & 1.00  & 1.00  & 259-2* & 0.98  & 0.95  & 0.96  \\
    237   & 1.00  & 1.00  & 1.00  & 259-4* & 0.98  & 0.95  & 0.96  \\
    238   & 0.99  & 0.99  & 0.99  & 259-6* & 0.98  & 0.95  & 0.96  \\
    239   & 0.97  & 1.00  & 0.98  & 259-8* & 0.98  & 0.95  & 0.96  \\
    240   & 0.97  & 1.00  & 0.98  & 260*   & 0.94  & 0.55  & 0.69  \\
    241   & 1.00  & 1.00  & 1.00  & 260-2* & 0.96  & 0.93  & 0.94  \\
    246   & 0.97  & 1.00  & 0.98  & 260-4* & 0.93  & 0.93  & 0.93  \\
    247   & 0.94  & 0.97  & 0.95  & 260-6* & 0.96  & 0.79  & 0.87  \\
    \cline{1-4}
    248*   & 1.00  & 0.81  & 0.90  & 260-8* & 0.88  & 0.72  & 0.79  \\
    248-2* & 1.00  & 0.95  & 0.97  & 261*   & 0.67  & 0.48  & 0.56  \\
    248-4* & 1.00  & 0.92  & 0.96  & 261-2* & 0.88  & 0.91  & 0.89  \\
    248-6* & 1.00  & 0.88  & 0.94  & 261-4* & 0.88  & 0.91  & 0.89  \\
    248-8* & 0.98  & 0.85  & 0.91  & 261-6* & 0.88  & 0.91  & 0.89  \\
    249*   & 0.83  & 0.66  & 0.74  & 261-8* & 0.88  & 0.91  & 0.89  \\
    249-2* & 0.98  & 0.95  & 0.96  & 262*   & 0.00  & 0.00  & 0.00  \\
    249-4* & 0.98  & 0.91  & 0.94  & 262-2* & 1.00  & 0.79  & 0.88  \\
    249-6* & 0.98  & 0.87  & 0.92  & 262-4* & 1.00  & 0.61  & 0.76  \\
    249-8* & 0.95  & 0.82  & 0.88  & 262-6* & 1.00  & 0.42  & 0.59  \\
    250*   & 0.90  & 0.58  & 0.71  & 262-8* & 1.00  & 0.21  & 0.35  \\
    250-2* & 1.00  & 1.00  & 1.00  & 265*   & 0.80  & 0.14  & 0.24  \\
    250-4* & 1.00  & 1.00  & 1.00  & 266*   & 0.30  & 0.09  & 0.14  \\
    \cline{5-8}
    250-6* & 1.00  & 1.00  & 1.00  & 301   & 0.94  & 0.82  & 0.88  \\
    250-8* & 1.00  & 0.88  & 0.94  & 302   & 0.89  & 0.65  & 0.75  \\
          &       &       &       & 303   & 0.65  & 0.71  & 0.68  \\
          &       &       &       & 304   & 0.95  & 0.97  & 0.96  \\
    \hline
    \end{tabular}%
\end{table*}%

\begin{sidewaystable*}[htbp]
	\centering

	\caption{Overall matching results of 15 systems on benchmark 2008}
		\label{omr_b2008}
	\begin{tabular}{llllllllllllllll}
		\hline
		\multicolumn{1}{c}{} & \multicolumn{3}{c}{SPIDER} & \multicolumn{3}{c}{ASMOV} & \multicolumn{3}{c}{DSSim} & \multicolumn{3}{c}{Anchor-Flood} & \multicolumn{3}{c}{Lily}  \\
		\cline{2-16}    \multicolumn{1}{c}{} & \tiny{Precision} & \tiny{Recall} & \tiny{F1-measure} & \tiny{Precision} & \tiny{Recall} & \tiny{F1-measure} & \tiny{Precision} & \tiny{Recall} & \tiny{F1-measure} & \tiny{Precision} & \tiny{Recall} & \tiny{F1-measure} & \multicolumn{1}{c}{\tiny{Precision}} & \multicolumn{1}{c}{\tiny{Recall}} & \multicolumn{1}{c}{\tiny{F1-measure}} \\
		\hline
		\multicolumn{1}{c}{101-104} & 0.99  & 0.99 & 0.99  & \textbf{1.00 } & \textbf{1.00 } & \textbf{1.00 } & \textbf{1.00 } & \textbf{1.00 } & \textbf{1.00 } & \textbf{1.00 } & \textbf{1.00 } & \textbf{1.00 } & \multicolumn{1}{c}{\textbf{1.00 }} & \multicolumn{1}{c}{\textbf{1.00 }} & \multicolumn{1}{c}{\textbf{1.00 }} \\
		\multicolumn{1}{c}{201-210} & 0.92  & 0.65  & 0.76  & 0.98  & \textbf{0.96 } & \textbf{0.97 } & 0.96  & 0.75  & 0.84  & 0.97  & 0.80  & 0.87  & \multicolumn{1}{c}{\textbf{1.00 }} & \multicolumn{1}{c}{0.95 } & \multicolumn{1}{c}{\textbf{0.97 }} \\
		\multicolumn{1}{c}{221-247} & 0.97  & 0.98 & 0.97  & 0.99  & \textbf{1.00 } & 0.99  & 0.99  & \textbf{1.00 } & 0.99  & 0.99  & \textbf{1.00 } & 0.99  & \multicolumn{1}{c}{0.99 } & \multicolumn{1}{c}{\textbf{1.00 }} & \multicolumn{1}{c}{0.99 } \\
		\multicolumn{1}{c}{248-266} & 0.76  & 0.48  & 0.59  & 0.90  & 0.73  & 0.80  & 0.90  & 0.51  & 0.65  & 0.89  & 0.56  & 0.69  & \multicolumn{1}{c}{\textbf{0.92 }} & \multicolumn{1}{c}{\textbf{0.76 }} & \multicolumn{1}{c}{\textbf{0.83 }} \\
		\multicolumn{1}{c}{301-304} & 0.19  & 0.80 & 0.31  & 0.78  & 0.75  & 0.76  & 0.89  & 0.69  & 0.78  & 0.71  & 0.61  & 0.65  & \multicolumn{1}{c}{0.86 } & \multicolumn{1}{c}{0.79 } & \multicolumn{1}{c}{0.82 } \\
		\hline
		\multicolumn{1}{c}{Avg.} & 0.80  & 0.61  & 0.66  & 0.93  & 0.82  & 0.85  & 0.93  & 0.65  & 0.73  & 0.92  & 0.69  & 0.76  & \multicolumn{1}{c}{\textbf{0.94 }} & \multicolumn{1}{c}{\textbf{0.84 }} & \multicolumn{1}{c}{\textbf{0.88 }} \\
		\hline
		\hline
		\multicolumn{1}{c}{} & \multicolumn{3}{c}{AROMA} & \multicolumn{3}{c}{CIDER} & \multicolumn{3}{c}{GeRoMe} & \multicolumn{3}{c}{MapPSO} & \multicolumn{3}{c}{RiMOM}  \\
		\cline{2-16}    \multicolumn{1}{c}{} & \tiny{Precision} & Recall & \tiny{F1-measure} & \tiny{Precision} & \tiny{Recall} & \tiny{F1-measure} & \tiny{Precision} & \tiny{Recall} & \tiny{F1-measure} & \tiny{Precision} & \tiny{Recall} & \tiny{F1-measure} & \multicolumn{1}{c}{\tiny{Precision}} & \multicolumn{1}{c}{\tiny{Recall}} & \multicolumn{1}{c}{\tiny{F1-measure}} \\
		\hline
		\multicolumn{1}{c}{101-104} & \textbf{1.00 } & \textbf{1.00 } & \textbf{1.00 } & 0.99  & 0.99  & 0.99  & 0.96  & 0.79  & 0.87  & 0.92  & \textbf{1.00 } & 0.96  & \multicolumn{1}{c}{\textbf{1.00 }} & \multicolumn{1}{c}{\textbf{1.00 }} & \multicolumn{1}{c}{1.00 } \\
		\multicolumn{1}{c}{201-210} & 0.98  & 0.86  & 0.91  & 0.92  & 0.65  & 0.76  & 0.85  & 0.69  & 0.76  & 0.46  & 0.50  & 0.47  & \multicolumn{1}{c}{0.99 } & \multicolumn{1}{c}{0.95 } & \multicolumn{1}{c}{\textbf{0.97 }} \\
		\multicolumn{1}{c}{221-247} & 0.97  & 0.98  & 0.97  & 0.97  & 0.98  & 0.97  & 0.92  & 0.73  & 0.81  & 0.86  & 0.96  & 0.91  & \multicolumn{1}{c}{\textbf{1.00 }} & \multicolumn{1}{c}{\textbf{1.00 }} & \multicolumn{1}{c}{\textbf{1.00 }} \\
		\multicolumn{1}{c}{248-266} & 0.83  & 0.54  & 0.66  & 0.76  & 0.48  & 0.59  & 0.51  & 0.49  & 0.50  & 0.43  & 0.41  & 0.42  & \multicolumn{1}{c}{\textbf{0.92 }} & \multicolumn{1}{c}{0.70 } & \multicolumn{1}{c}{0.79 } \\
		\multicolumn{1}{c}{301-304} & 0.81  & 0.68  & 0.74  & 0.90  & 0.73  & 0.81  & 0.46  & 0.37  & 0.41  & 0.22  & 0.21  & 0.21  & \multicolumn{1}{c}{0.79 } & \multicolumn{1}{c}{0.80 } & \multicolumn{1}{c}{0.79 } \\
		\hline
		\multicolumn{1}{c}{Avg.} & 0.88  & 0.68  & 0.75  & 0.83  & 0.61  & 0.68  & 0.64  & 0.57  & 0.58  & 0.51  & 0.53  & 0.51  & \multicolumn{1}{c}{\textbf{0.94 }} & \multicolumn{1}{c}{0.80 } & \multicolumn{1}{c}{0.85 } \\
		\hline
		\hline
		\multicolumn{1}{l}{} & \multicolumn{3}{c}{SAMBO} & \multicolumn{3}{c}{SAMBOdtf}  & \multicolumn{3}{c}{TaxoMap} & \multicolumn{3}{c}{AML} & \multicolumn{3}{c}{LogMap} \\
		\cline{2-16}    \multicolumn{1}{l}{} & \tiny{Precision} & \tiny{Recall} & \tiny{F1-measure} & \tiny{Precision} & \tiny{Recall} & \tiny{F1-measure} & \tiny{Precision} & \tiny{Recall} & \tiny{F1-measure} & \tiny{Precision} & \tiny{Recall} & \tiny{F1-measure} & \tiny{Precision} & \tiny{Recall} & \tiny{F1-measure}          \\
		\hline
		\multicolumn{1}{c}{101-104} & \textbf{1.00 } & \textbf{1.00 } & \textbf{1.00 } & \textbf{1.00 } & \textbf{1.00 } & \textbf{1.00 }   & \textbf{1.00 } & 0.34  & 0.51  &  0.75 & 0.72  & 0.73  & 0.68  & 0.69  & 0.69\\
		\multicolumn{1}{c}{201-210} & 0.98  & 0.52  & 0.68  & 0.98  & 0.55  & 0.71    & 0.67  & 0.13  & 0.22  & \textbf{1.00}	& 0.43	& 0.53 & 0.44  & 0.21  & 0.27 \\
		\multicolumn{1}{c}{221-247} & 0.99  & 0.99  & 0.99  & 0.99  & 0.99  & 0.99   & 0.99  & 0.62  & 0.76  &  0.88	& 0.86	&0.86  & 0.84  & 0.97  & 0.90\\
		\multicolumn{1}{c}{248-266} & 0.87  & 0.44  & 0.59  & 0.87  & 0.44  & 0.59   & 0.67  & 0.23  & 0.34  &   0.81	&0.41	&0.50 & 0.50  & 0.33  & 0.38  \\
		\multicolumn{1}{c}{301-304} & \textbf{0.95 } & 0.79  & \textbf{0.86 } & 0.92  & 0.79  & 0.85    & 0.70  & 0.19  & 0.30  &0.85	&0.39 &0.51& 0.93  & 0.38  & 0.52\\
		\hline
		\multicolumn{1}{c}{Avg.} & 0.91  & 0.58  & 0.65  & 0.91  & 0.58  & 0.66    & 0.73  & 0.28  & 0.37  &0.85	&0.49	&0.57  & 0.57  & 0.43  & 0.46\\
		\hline
	\end{tabular}%
	\label{tab:addlabel}%
\end{sidewaystable*}%

\begin{sidewaystable}[htbp]
  \centering

  \caption{Overall matching results of 15 systems on benchmark 2009}
    \label{omr_b2009}
    \begin{tabular}{lllllllllllllllll}
    \hline
    \multicolumn{1}{c}{} & \multicolumn{3}{c}{enda} & \multicolumn{3}{c}{ASMOV} & \multicolumn{3}{c}{DSSim} & \multicolumn{3}{c}{Anchor-Flood} & \multicolumn{3}{c}{Lily} \\
\cline{2-16}    \multicolumn{1}{c}{} & \multicolumn{1}{c}{\tiny{Precision}} & \multicolumn{1}{c}{\tiny{Recall}} & \multicolumn{1}{c}{\tiny{F1-measure}} & \multicolumn{1}{c}{\tiny{Precision}} & \multicolumn{1}{c}{\tiny{Recall}} & \multicolumn{1}{c}{\tiny{F1-measure}} & \multicolumn{1}{c}{\tiny{Precision}} & \multicolumn{1}{c}{\tiny{Recall}} & \multicolumn{1}{c}{\tiny{F1-measure}} & \multicolumn{1}{c}{\tiny{Precision}} & \multicolumn{1}{c}{\tiny{Recall}} & \multicolumn{1}{c}{\tiny{F1-measure}} & \multicolumn{1}{c}{\tiny{Precision}} & \multicolumn{1}{c}{\tiny{Recall}} & \multicolumn{1}{c}{\tiny{F1-measure}} \\
    \hline
    \multicolumn{1}{c}{101-104} & 0.96  & \textbf{1.00 } & 0.98  & \textbf{1.00 } & \textbf{1.00 } & \textbf{1.00 } & \textbf{1.00 } & \textbf{1.00 } & \textbf{1.00 } & \textbf{1.00 } & \textbf{1.00 } & \textbf{1.00 } & \textbf{1.00 } & \textbf{1.00 } & \textbf{1.00 } \\
    \multicolumn{1}{c}{201-210} & 0.50  & 0.52  & 0.51  & 0.98  & \textbf{0.96 } & \textbf{0.97 } & 0.95  & 0.70  & 0.78  & 0.98  & 0.89  & 0.93  & \textbf{0.99 } & 0.95  & \textbf{0.97 } \\
    \multicolumn{1}{c}{221-247} & 0.67  & \textbf{1.00 } & 0.76  & \textbf{0.99 } & \textbf{1.00 } & \textbf{0.99 } & \textbf{0.99 } & \textbf{1.00 } & \textbf{1.00 } & \textbf{0.99 } & \textbf{1.00 } & \textbf{0.99 } & \textbf{0.99 } & \textbf{1.00 } & \textbf{0.99 } \\
    \multicolumn{1}{c}{248-266} & 0.31  & 0.45  & 0.35  & 0.90  & 0.73  & 0.79  & 0.91  & 0.52  & 0.62  & 0.96  & 0.70  & 0.77  & \textbf{0.93 } & \textbf{0.76 } & \textbf{0.82 } \\
    \multicolumn{1}{c}{301-304} & 0.47  & 0.80  & 0.59  & 0.80  & \textbf{0.80 } & 0.79  & \textbf{0.92 } & 0.65  & 0.74  & 0.90  & 0.78  & 0.83  & 0.83  & 0.79  & 0.80  \\
    \hline
    \multicolumn{1}{c}{Avg.} & 0.42  & 0.58  & 0.47  & 0.93  & 0.82  & 0.86  & 0.93  & 0.64  & 0.72  & \textbf{0.97 } & 0.79  & 0.84  & 0.95  & \textbf{0.84 } & \textbf{0.88 } \\
    \hline
    \hline
    \multicolumn{1}{c}{} & \multicolumn{3}{c}{AROMA} & \multicolumn{3}{c}{AgrMaker} & \multicolumn{3}{c}{GeRoMe} & \multicolumn{3}{c}{MapPSO} & \multicolumn{3}{c}{RiMOM} \\
\cline{2-16}    \multicolumn{1}{c}{} & \multicolumn{1}{c}{\tiny{Precision}} & \multicolumn{1}{c}{\tiny{Recall}} & \multicolumn{1}{c}{\tiny{F1-measure}} & \multicolumn{1}{c}{\tiny{Precision}} & \multicolumn{1}{c}{\tiny{Recall}} & \multicolumn{1}{c}{\tiny{F1-measure}} & \multicolumn{1}{c}{\tiny{Precision}} & \multicolumn{1}{c}{\tiny{Recall}} & \multicolumn{1}{c}{\tiny{F1-measure}} & \multicolumn{1}{c}{\tiny{Precision}} & \multicolumn{1}{c}{\tiny{Recall}} & \multicolumn{1}{c}{\tiny{F1-measure}} & \multicolumn{1}{c}{\tiny{Precision}} & \multicolumn{1}{c}{\tiny{Recall}} & \multicolumn{1}{c}{\tiny{F1-measure}} \\
    \hline
    \multicolumn{1}{c}{101-104} & \textbf{1.00 } & \textbf{1.00 } & \textbf{1.00 } & 0.98  & 0.98  & 0.98  & \textbf{1.00 } & \textbf{1.00 } & \textbf{1.00 } & \textbf{1.00 } & \textbf{1.00 } & \textbf{1.00 } & \textbf{1.00 } & \textbf{1.00 } & \textbf{1.00 } \\
    \multicolumn{1}{c}{201-210} & 0.98  & 0.85  & 0.90  & \textbf{0.99 } & 0.70  & 0.76  & 0.92  & 0.86  & 0.88  & 0.90  & 0.90  & 0.90  & \textbf{0.99 } & 0.95  & \textbf{0.97 } \\
    \multicolumn{1}{c}{221-247} & 0.97  & 0.98  & 0.97  & 0.98  & 0.99  & 0.98  & \textbf{0.99 } & \textbf{1.00 } & \textbf{0.99 } & 0.97  & 0.98  & 0.97  & \textbf{0.99 } & \textbf{1.00 } & \textbf{0.99 } \\
    \multicolumn{1}{c}{248-266} & 0.84  & 0.52  & 0.61  & 0.87  & 0.45  & 0.54  & 0.81  & 0.58  & 0.66  & 0.39  & 0.39  & 0.39  & 0.84  & 0.69  & 0.74  \\
    \multicolumn{1}{c}{301-304} & 0.83  & 0.76  & 0.79  & \textbf{0.92 } & 0.77  & \textbf{0.83 } & 0.71  & 0.58  & 0.63  & 0.27  & 0.24  & 0.25  & 0.80  & \textbf{0.80 } & 0.80  \\
    \hline
    \multicolumn{1}{c}{Avg.} & 0.89  & 0.67  & 0.73  & 0.92  & 0.60  & 0.67  & 0.86  & 0.71  & 0.76  & 0.58  & 0.58  & 0.58  & 0.89  & 0.79  & 0.83  \\
    \hline
    \hline
    \multicolumn{1}{l}{} & \multicolumn{3}{c}{kosimap} & \multicolumn{3}{c}{SOBOM} & \multicolumn{3}{c}{TaxoMap} & \multicolumn{3}{c}{AML}  & \multicolumn{3}{c}{LogMap} \\
\cline{2-16}    \multicolumn{1}{l}{} & \multicolumn{1}{c}{\tiny{Precision}} & \multicolumn{1}{c}{\tiny{Recall}} & \multicolumn{1}{c}{\tiny{F1-measure}} & \multicolumn{1}{c}{\tiny{Precision}} & \multicolumn{1}{c}{\tiny{Recall}} & \multicolumn{1}{c}{\tiny{F1-measure}} & \multicolumn{1}{c}{\tiny{Precision}} & \multicolumn{1}{c}{\tiny{Recall}} & \multicolumn{1}{c}{\tiny{F1-measure}} & \multicolumn{1}{c}{\tiny{Precision}} & \multicolumn{1}{c}{\tiny{Recall}} & \multicolumn{1}{c}{\tiny{F1-measure}}&\multicolumn{1}{c}{\tiny{Precision}} & \multicolumn{1}{c}{\tiny{Recall}} & \multicolumn{1}{c}{\tiny{F1-measure}}\\
    \hline
    \multicolumn{1}{c}{101-104} & 0.99  & 0.99  & 0.99  & 0.98  & 0.97  & 0.98  & \textbf{1.00 } & 0.34  & 0.51  &  \textbf{1.00}     & 0.96   & 0.98   &   0.91    & 0.92      & 0.92 \\
    \multicolumn{1}{c}{201-210} & 0.88  & 0.61  & 0.69  & 0.72  & 0.35  & 0.43  & 0.65  & 0.17  & 0.26  &  0.89  &  0.36  &  0.46  &  0.50     & 0.26       & 0.32  \\
    \multicolumn{1}{c}{221-247} & 0.96  & 0.97  & 0.96  & 0.96  & 0.96  & 0.96  & 0.97  & 0.63  & 0.71  &  \textbf{0.99}     & 0.96      & 0.98      &   0.73    & 0.76      & 0.73 \\
    \multicolumn{1}{c}{248-266} & 0.86  & 0.45  & 0.54  & 0.78  & 0.33  & 0.44  & 0.64  & 0.30  & 0.37  &  0.80     & 0.40      & 0.49      &   0.52    &  0.36     & 0.40 \\
    \multicolumn{1}{c}{301-304} & 0.69  & 0.48  & 0.56  & 0.86  & 0.52  & 0.60  & 0.77  & 0.31  & 0.43  &  0.85     & 0.39      & 0.51      &   0.93    &  0.39     & 0.53 \\
    \hline
    \multicolumn{1}{c}{Avg.} & 0.88  & 0.58  & 0.65  & 0.81  & 0.46  & 0.54  & 0.71  & 0.33  & 0.41  &  0.85  &  0.50  & 0.58  & 0.58      &  0.43     & 0.47\\
    \hline
    \end{tabular}%
  \label{tab:addlabel}%
\end{sidewaystable}%

Our matching method for WIOs has been implemented in our ontology matching system Lily. 
Lily first checks whether an ontology is weak informative, then uses similarity propagation model to match weak informative ontologies and uses string-based matcher based on semantic graphs (in Section 9) to match other ontologies. 

Table~\ref{gmr_b2008} presents detailed matching results on benchmark2008 by Lily. 
Matching task numbers with star are weak informative matching tasks. 
Lily performs very well on the informative ontology matching tasks. For group 101-104 and 221-247, matching results are perfect and average precision, recall and F1-measure are all 0.99. For the real-world group 301-304, Lily also obtains good results with average 0.86 precision, 0.79 recall and 0.82 F1-measure.
For most weak informative ontologies matching tasks in group 201-210 and 248-266, Lily also gets good results. 
We notice that matching results on 254, 257, 261, 262, 262-6, 262-8, 265 and 266 have less than 0.6 F1-measure. 
The reason is that not only the ontology structures are suppressed, but also almost all meaningful literal information is removed. 

Table~\ref{omr_b2008} shows results of 15 matching systems on benchmark2008 \cite{OAEI08, Wang08}, which is also divided into five groups: 101-104, 201-210, 221-247, 248-266 and 301-304. Here the average results on each group are presented in a row, and the last row shows the overall average results for all matching tasks. 
(1) Group 101-104 contains some simple matching tasks, except GeRoMe and TaxoMap, other 12 systems obtain good results, and 8 systems including Lily have 1.00 F1-measure. 
(2) For group 201-210, since some ontologies are weak informative, some systems do not perform well.  
F1-measure of 6 systems is higher than 0.80. 
F1-measure of 4 systems is higher than 0.90. 
The best three systems: ASMOV, RiMOM and Lily get 0.97 F1-measure.
(3) For group 221-247, 11 systems perform very well and have F1-measure of higher than 0.90. 
RiMOM has the perfect result with 1.00 F1-measure. Lily has similar results with 0.99 F1-measure.
(4) For group 248-266, most systems cannot deal with these tasks efficiently. 
Only 3 systems have higher than 0.70 F1-measure, and they are ASMOV, Lily and RiMOM. 
Only Lily and ASMOV have 0.80 and 0.82 F1-measure, respectively. 
Lily has 0.92 precision and 0.76 recall, which are the best results in all systems. 
Lily performs best on this difficult group and obtains better results than most systems. 
(5) For group 301-304, not all systems can perform well on the real-world matching tasks, and only 4 systems have higher than 0.8 F1-measure. 
Best F1-measure is 0.86 by SAMBO. Lily has 0.82 F1-measure.
For the overall average results, only 3 systems, ASMOV, RiMOM and Lily, have higher than 0.80 F1-measure. 
That dues to good performance on weak informative ontology matching tasks. 
Lily has average 0.88 F1-measure, 0.94 precision and 0.84 recall, all of them are the best in all systems.

Table~\ref{omr_b2009} shows results of 15 systems on benchmark2009 \cite{OAEI09,Wang09OM}. Lily is also one of the best systems. 
Especially for group 248-266, Lily still obtains best results. 
For the overall average results, 4 systems have average higher than 0.8 F1-measure: ASMOV, Lily, Anchor-Flood and RiMOM. 
Lily has the best F1-measure of 0.88 and best recall of 0.84. 
Lily also has 0.95 precision, only lower than 0.97 precision by Anchor-Flood.


\section{Related Work}

Many ontology matching techniques have been  proposed. 
Some researchers have given very comprehensive surveys for this open problem \cite{Kalfoglou03, Shvaiko05, Rahm01,  Euzenat2013, Shvaiko2012, Cerdeira2015, Ochieng2018}. 
In recent years, this problem still attracts the attentions of researchers, and new ontology matching methods are proposed.
Zhao et al. proposed a method based on formal concept analysis to identify and validate mappings across ontologies \cite{songmao18}.
Faria et al. discussed the methodology for automatically selecting background knowledge sources for any given ontologies to match \cite{faria14}.
The simulated annealing, which is a evolutionary algorithm, was also used to find the mappings between two given ontologies while no ground truth is available \cite{SANOM20}. 
However, these works ignore the weak informative ontologies in a lot of practical applications. 
This paper mainly focuses on the weak informative ontology matching problem. 
Structure-based method is a feasible way to deal with this special matching problem.
For the reason that the ontology graph topology cannot represent the semantics reasonably, methods using classical graph matching algorithm cannot work and has unsatisfied matching performance. 
Therefore, the method based on similarity propagation idea is the practical way to solve the problem.

Blondel et al. proposed a iteration equation for measuring the similarity between directed graphs \cite{Blondel04}. 
It is based on the $Hub-Authority$ idea, which is a kind of similarity propagation. 
Leicht et al. proposed a more general measurement for the vertex similarity in network \cite {Leicht06}, they also pointed out that Blondel's method is a special case of their method.
However, the two graph matching methods can only calculate the vertex similarity in graph, but cannot deal with the edge similarity.
To overcome this shortcoming, Hu et al. used the bipartite graph to represent the ontology graph, and then they modified Blondel's method to handle ontology graph matching problem \cite{HU05}. 
This method is implemented in Falcon-AO \cite{Falcon2008}. 
Inspired by these work, Tous and Delgado first represented ontology graphs as the vector space, then used Blondel's graph matching algorithm to matching weak informative ontologies \cite {Tous06} and obtained close performance to Falcon-AO.

Similarity flooding \cite {Melnik02} is the most popular algorithm inspired by the similarity propagation,
but it cannot be directly used for ontology matching (see the reasons we discussed in previous sections). 
However, some weak informative ontology matching methods are based on similarity flooding.

Noy and Musen proposed AnchorPrompt algorithm \cite{Noy03} to find alignment between concepts by analyzing the ontology graph structure. 
The principle of AnchorPrompt is that if two pairs of elements from different ontologies are similar and there are paths connecting the elements, then elements in those paths are often similar as well.
Therefore, it is also a variation of the similarity flooding idea. 
However, this approach does not work well when ontologies are constructed in different ways. 
For example, one ontology has a deep concept hierarchy with many inter-linked concepts and another ontology is a shallow one where the hierarchy has only a few levels.

In ontology mapping system RiMOM \cite {RiMOM09}, Li and Tang et al. used three propagation strategies for structure matching: (1) propagation between concepts; (2) propagation between relations; (3) propagation between concepts and relations. 
This method employs ontological structure information to design  propagation for matching ontologies. 
It is a very successful variation of similarity flooding for ontology matching.

In ASMOV \cite{ASMOV2009}, Jean-Mary et al. used two structural similarities: a relational similarity and an internal similarity, to discover alignments. 
The relational similarity is computed by combining the similarities between parents and children. 
The internal similarity is calculated by considering domain and range of properties, property restrictions associated to a concept and other structure information. 
These structural similarities are calculated iteratively, and also use similarities propagated by neighbors. 
Therefore, it can be seen as a special case of similarity flooding.

Seddiqui and Aono proposed Anchor-Flood algorithm \cite{AFlood2009}. 
This algorithm first uses few initial correspondences and gradually explored concepts by collecting concept neighbors and other graph data structure to output a set of correspondences between concepts and properties in semantically connected subsets. 
The process runs iteratively until the algorithm satisfies the condition: \emph{either all the collected concepts are explored, or no new aligned pair is found}. 
This method is also inspired by the idea of similarity flooding. 

In AgreementMaker \cite{AMaker2009}, Cruz et al.used a structure-based matcher called DSI ( Descendant's Similarity Inheritance).
DSI is based on the heuristic rule: if two nodes are matched with high similarity, then the similarity between the descendants of those nodes should increase. 
It is also a kind of variation of similarity flooding. 
DSI matcher is also used by another system CSA \cite{CSA2011}. 

In SAMBO \cite{SAMBO2006}, Lambrix and Tan proposed a structural matcher, which was an iterative algorithm based on the \emph{is-a} and \emph{part-of} hierarchies of ontologies. 
The algorithm is based on the intuition: if two concepts are in similar positions with respect to \emph{is-a} or \emph{part-of} hierarchies  and are relative to already aligned concepts, then they are likely to be similar as well.

In GeRoMe \cite{GeRoMe2007, GeRoMe2009}, Quix et al. also found that direct Similarity Flooding had no positive effect on the matching quality. 
Meanwhile,  this matching system designed structural similarity called children and parent matchers, which propagate the similarity of elements up and down in the class hierarchy. 

In PRIOR+ \cite{Mao2010}, Mao et al. did not directly use similarity flooding, but they proposed an important similarity called profile similarity. 
This method first generates a profile for each element using label, comments and other related literal information of the element, then the profile of the element’s ancestors, descendants and siblings will be passed to that of the element with different weights. 
Finally, the cosine similarity between the profiles of two elements is calculated in a vector space model. 
We believe this is also a kind of variation of similarity flooding, because the profile propagation is a kind of similarity propagation.

In LogMap \cite{LogMap2011, LogMap2018}, Jiménez-Ruiz et al. realized a similar propagation idea. First, LogMap computes an initial set of equivalence anchor mappings by intersecting the lexical indexes of each input ontology. Then these mappings will later serve as starting point for the further discovery of additional mappings. This mapping discovery strategy is based on a principle: if classes $C_1$ and $C_2$ are correctly mapped, then the classes semantically related to $C_1$ in $O_1$ are likely to be mapped to those semantically related to $C_2$ in $O_2$.

Similarity flooding idea is also used in DSSim \cite{DSSim2010} and YAM++ \cite{YAM2011}.
YAM++ proposed the confidence propagation based on similarity flooding \cite{YAM2016}. 
The intuition behind the confidence propagation is that  the similarity of the two matched concepts will contribute in some degree of confidence to the similarity of their relatives along the same path of semantic relations.

Since some weak informative ontologies in OAEI benchmark datasets can be seen as different versions of the same ontology, 
CODI \cite{CODI2010} first matches different ontology versions, then refines the matching results by markov logic framework \cite{CODI2011}. 
However, this method also can be used in matching tasks not generated by different versions.

Similar problem also arises in schema matching in database research field. For example, the column names in the schemas and the data in the columns are opaque or very difficult to interpret. 
Kang and Naughton presented two-step technique that works even in the presence of opaque column names and data values \cite{Kang2003}. 
They constructed a dependency graph using mutual information, then found matching node pairs in dependency graphs by running a graph matching algorithm. 
However, it is an instance-based matcher and cannot solve the concept matching or property matching in ontologies.

Compared to current works, our method has some features: 
(1) It is the first general similarity propagation framework for ontology matching instead of some heuristic propagation rules, and it can discover alignments of concepts, properties and instances.  
(2) It uses semantic subgraphs to capture precise meanings of ontology elements, then constructs semantic description documents, finally, designs a matcher to provide few credible correspondences as seeds for similarity propagation model. 
(3) It utilizes the strong constraint condition and semantic subgraphs to restrict the range of similarity propagation, that can avoid the meaningless propagations. 
Meanwhile, our method also has  limitations: 
(1) First, extracting semantic subgraphs for all elements is a time-consuming process, in which solving the linear equations in the circuit model is the time bottleneck. Fortunately, this limitation can be solved by parallel computing.
(2) Second, there are more parameters (such as semantic graph size) and settings (such as propagation strategy) in our method, it needs more experiences to tune the parameters and select reasonable settings. 
In our ongoing work, we are designing the automatic tuning method to help users to select optimized parameters and settings.

\section{Conclusion}

This paper proposes a method for matching weak informative ontologies. 
The success of our method is due to two techniques:  semantic subgraphs and a new similarity propagation model.
Semantic subgraphs can precisely describe ontology elements with limited triples. 
The similarity propagation model is based on the strong constrained condition, and it is reasonable for handling ontology matching. 
Some useful propagation strategies are also adopted to improve matching results.
Our method is used for discovering correspondences between concepts or properties. 
However, in recent years,  knowledge base such as knowledge graph \cite{KG2014} usually contains large scale instances, which would also be weak informative.
In the very near future, we will extend our method for matching large scale knowledge graphs.
In addition, except using WordNet to define weak informative ontology, we do not use any external knowledge or data in ontology matching.
In the future, we will enhance the matcher based on semantic subgraphs using word2vec \cite{Mikolov2013} and knowledge graph embedding \cite{Bordes2013}, which are trained on external corpus and knowledge graphs respectively.   
That would provide more credible seeds for similarity propagation.  




\bibliographystyle{elsarticle-num}
\bibliography{MWOBib}

\begin{thebibliography}{10}
\expandafter\ifx\csname url\endcsname\relax
  \def\url#1{\texttt{#1}}\fi
\expandafter\ifx\csname urlprefix\endcsname\relax\def\urlprefix{URL }\fi
\expandafter\ifx\csname href\endcsname\relax
  \def\href#1#2{#2} \def\path#1{#1}\fi

\bibitem{Doan02}
A.~Doan, J.~Madhavan, P.~Domingos, A.~Halevy, Learning to map between
  ontologies on the semantic web, in: Proceedings of the 11th international
  conference on World Wide Web (WWW2002)., 2002.

\bibitem{Aumueller05}
D.~Aumueller, H.-H. Do, S.~Massmann, E.~Rahm, Schema and ontology matching with
  coma++, in: Proceedings of the 2005 ACM SIGMOD international conference on
  management of data., 2005.

\bibitem{Rahm01}
E.~Rahm, P.~A. Bernstein, A survey of approaches to automatic schema matching,
  The VLDB Journal 10 (2001) 334--350.

\bibitem{Kalfoglou03}
Y.~Kalfoglou, M.~Schorlemmer, Ontology mapping: the state of the art, The
  Knowledge Engineering Review 18(1) (2003) 1--31.

\bibitem{Shvaiko05}
P.~Shvaiko, J.~Euzenat, A survey of schema-based matching approaches, Journal
  on Data Semantics 4 (2005) 146--171.

\bibitem{Shvaiko2012}
P.~Shvaiko, J.~Euzenat, Ontology matching: state of the art and future
  challenges, IEEE Transactions on Knowledge and Data Engineering 25(1) (2011)
  158--176.

\bibitem{Euzenat2013}
J.~Euzenat, P.~Shvaiko, Ontology matching (2nd edition), Heidelberg:
  Springer-Verlag, 2013.

\bibitem{Cerdeira2015}
L.~Otero-Cerdeira, F.~J. Rodríguez-Martínez, A.~Gómez-Rodríguez, Ontology
  matching: a literature review, Expert Systems with Applications 42(2) (2015)
  949--971.

\bibitem{Ochieng2018}
P.~Ochieng, S.~Kyanda, Large-scale ontology matching: state-of-the-art
  analysis, ACM Computing Surveys 51(4) (2018) 1--35.

\bibitem{Pfisterer2011}
D.~Pfisterer, K.~Römer, D.~Bimschas, et~al., Spitfire: toward a semantic web
  of things, IEEE Communications Magazine 49(11) (2011) 40--48.

\bibitem{Tang_Grid19}
Y.~Tang, T.~Liu, G.~Liu, et~al., Enhancement of power equipment management
  using knowledge graph, in: Proceedings of the IEEE Innovative Smart Grid
  Technologies-Asia . Chengdu, China, 2019.

\bibitem{Huang_JPES15}
Y.~Huang, X.~Zhou, Knowledge model for electric power big data based on
  ontology and semantic web, CSEE Journal of Power and Energy Systems 1(1)
  (2015) 19--27.

\bibitem{Bader_ESWC20}
S.~R. Bader, I.~Grangel-Gonzalez, P.~Nanjappa, et~al., A knowledge graph for
  industry 4.0, in: Proceedings of the 17th European Semantic Web Conference
  (ESWC2020). Heraklion, Crete, Greece, 2020.

\bibitem{Conte04}
D.~Conte, P.~Foggia, C.~Sansone, et~al., Thirty years of graph matching in
  pattern recognition, International Journal of Pattern Recognition and
  Artificial Intelligence 18(3) (2004) 265--298.

\bibitem{Falcon2008}
W.~Hu, Y.~Qu, Falcon-ao: A practical ontology matching system, Journal of Web
  Semantics (2008) 237--239.

\bibitem{RiMOM09}
J.~Li, J.~Tang, Y.~Li, et~al., Rimom: a dynamic multistrategy ontology
  alignment framework, IEEE Transactions on Knowledge and Data Engineering
  21(8) (2009) 1218--1232.

\bibitem{Melnik02}
S.~Melnik, H.~Garcia-Molina, E.~Rahm, Similarity flooding: a versatile graph
  matching algorithm and its application to schema matching, in: Proceeding of
  the 18th International Conference on Data Engineering (ICDE2002). San Jose,
  CA, USA, 2002.

\bibitem{Jeh02}
G.~Jeh, J.~Widom, Simrank: a measure of structural-context similarity, in:
  Proceedings of the 8th ACM SIGKDD International Conference on Knowledge
  Discovery and Data Mining (KDD2002). Edmonton, Alberta, Canada, 2002.

\bibitem{Blondel04}
V.~D. Blondel, A.~Gajardo, M.~Heymans, et~al., A measure of similarity between
  graph vertices: applications to synonym extraction and web searching, SIAM
  Review 46(4) (2004) 647--666.

\bibitem{DoanA2005}
A.~Doan, A.~Y. Halevy, Semantic integration research in the database community:
  a brief survey, AI magazine 26(1) (2005) 83--94.

\bibitem{Wang09ASWC}
P.~Wang, B.~Xu, An effective similarity propagation model for matching
  ontologies without sufficient or regular linguistic information, in:
  Proceedings of the 4th Asian Semantic Web Conference (ASWC2009). Shanghai,
  China, 2009.

\bibitem{OAEI08}
C.~Caraciolo, J.~Euzenat, L.~Hollink, et~al., Results of the ontology alignment
  evaluation initiative 2008, in: Proceedings of the 3rd ISWC Workshop on
  Ontology Matching (OM2008). Karlsruhe, Germany, 2008.

\bibitem{AML2013}
D.~Faria, C.~Pesquita, E.~Santos, et~al., The agreementmakerlight ontology
  matching system, in: Proceedings of the 12th On the Move to Meaningful
  Internet Systems (OTM2013), Graz, Austria, 2013.

\bibitem{AML2018}
D.~Faria, C.~Pesquita, B.~S. Balasubramani, et~al., Results of aml
  participation in oaei 2018, in: Proceedings of the 12th International
  Workshop on Ontology Matching (OM2018), Monterey, CA, USA, 2018.

\bibitem{LogMap2011}
E.~Jiménez-Ruiz, B.~C. Grau, Logmap: Logic-based and scalable ontology
  matching, in: Proceedings of the 10th International Semantic Web Conference
  (ISWC2011), Bonn, Germany, 2011.

\bibitem{LogMap2018}
E.~Jiménez-Ruiz, B.~C. Grau, V.~Cross, Logmap family participation in the oaei
  2018, in: Proceedings of the 12th International Workshop on Ontology Matching
  (OM2018), Monterey, CA, USA, 2018.

\bibitem{SOQA06}
P.~Ziegler, C.~Kiefer, C.~Sturm, K.~R. Dittrich, A.~Bernstein, Generic
  similarity detection in ontologies with the soqa-simpack toolkit, in:
  Proceedings of the ACM SIGMOD International Conference on Management of Data
  (SIGMOD2006). Chicago, Illinois, USA, 2006.

\bibitem{QOM04}
M.~Ehrig, S.~Staab, Qom - quick ontology mapping, in: Proceedings of the 3rd
  International Semantic Web Conference (ISWC2004). Hiroshima, Japan, 2004.

\bibitem{Noy03}
N.~F. Noy, M.~A. Musen, The prompt suite: interactive tools for ontology
  merging and mapping, International Journal of Human-Computer Studies 59(6)
  (2003) 983--1024.

\bibitem{HU05}
W.~Hu, N.~Jian, Y.~Qu, et~al., Gmo: a graph matching for ontologies, in:
  Proceedings of K-CAP Workshop on Integrating Ontologies. Banff, Alberta,
  Canada, 2005.

\bibitem{Leicht06}
E.~A. Leicht, P.~Holme, M.~E.~J. Newman, Vertex similarity in networks,
  Physical Review E 73.

\bibitem{Faloutsos2004}
C.~Faloutsos, K.~S. McCurley, A.~Tomkins, Fast discovery of connection
  subgraphs, in: Proceedings of the 10th ACM SIGKDD International Conference on
  Knowledge Discovery and Data Mining (KDD2004). Seattle, Washington, USA.,
  2004.

\bibitem{Qu2006}
Y.~Qu, W.~Hu, G.~Cheng, Constructing virtual documents for ontology matching,
  in: Proceedings of the 15th International Conference on World Wide Web
  (WWW2006). Edinburgh, Scotland, UK, 2006.

\bibitem{Wang08}
P.~Wang, B.~Xu, Lily: ontology alignment results for oaei 2008, in: Proceedings
  of the 3rd International Workshop on Ontology Matching (OM2008). Karlsruhe,
  Germany, 2008.

\bibitem{OAEI09}
J.~Euzenat, A.~Ferrara, L.~Hollink, et~al., Results of the ontology alignment
  evaluation initiative 2009, in: Proceedings of the 4th ISWC Workshop on
  Ontology Matching (OM2009). Trento, Italy, 2009.

\bibitem{Wang09OM}
P.~Wang, B.~Xu., Lily: ontology alignment results for oaei 2009, in:
  Proceedings of the 4th International Workshop on Ontology Matching (OM2009).
  Chantilly, VA, USA, 2009.

\bibitem{songmao18}
M.~Zhao, S.~Zhang, W.~Li, G.~Chen, Matching biomedical ontologies based on
  formal concept analysis, Journal of Biomedical Semantics 9(11) (2018) 1--27.

\bibitem{faria14}
D.~Faria, C.~Pesquita, E.~Santos, I.~F. Cruz, F.~M. Couto, Automatic background
  knowledge selection for matching biomedical ontologies, PLoS ONE 9(11) (2014)
  e111226.

\bibitem{SANOM20}
M.~Mohammadi, W.~Hofman, Y.-H. Tan, Sanom-hobbit: simulated annealing-based
  ontology matching on hobbit platform, The Knowledge Engineering Review 35(13)
  (2020) 1--13.

\bibitem{Tous06}
R.~Tous, J.~Delgado, A vector space model for semantic similarity calculation
  and owl ontology alignment, in: Proceedings of the 17th International
  Conference on Database and Expert Systems Applications (DEXA2006). Krakow,
  Poland, 2006.

\bibitem{ASMOV2009}
Y.~R. Jean-Mary, E.~P. Shironoshita, M.~R. Kabuka, Ontology matching with
  semantic verification, Journal of Web Semantics 7(3) (2009) 235--251.

\bibitem{AFlood2009}
M.~H. Seddiqui, M.~Aono, An efficient and scalable algorithm for segmented
  alignment of ontologies of arbitrary size, Journal of Web Semantics 7(4)
  (2009) 344--356.

\bibitem{AMaker2009}
I.~F. Cruz, F.~P. Antonelli, C.~Stroe, Agreementmaker: efficient matching for
  large real-world schemas and ontologies, PVLDB 2(2) (2009) 1586--1589.

\bibitem{CSA2011}
Q.-V. Tran, R.~Ichise, B.-Q. Ho, Cluster-based similarity aggregation for
  ontology matching, in: Proceedings of the 6th International Workshop on
  Ontology Matching (OM2011). Bonn, Germany, 2011.

\bibitem{SAMBO2006}
P.~Lambrix, H.~Tan, Sambo - a system for aligning and merging biomedical
  ontologies, Journal of Web Semantics 4 (2006) 196--206.

\bibitem{GeRoMe2007}
D.~Kensche, C.~Quix, L.~X, et~al., Geromesuite: A system for holistic generic
  model management, in: Proceedings of the 33rd International Conference on
  Very Large Data Bases (VLDB2007). Vienna, Austria, 2007.

\bibitem{GeRoMe2009}
C.~Quix, S.~Geisler, D.~Kensche, et~al., Results of geromesuite for oaei 2009,
  in: Proceedings of the 4th International Workshop on Ontology Matching
  (OM2009). Chantilly, VA, USA, 2009.

\bibitem{Mao2010}
M.~Mao, Y.~Peng, M.~Spring, An adaptive ontology mapping approach with neural
  network based constraint satisfaction, Journal of Web Semantics 8(1) (2010)
  14--25.

\bibitem{DSSim2010}
M.~Nagy, M.~Vargas-Vera, Towards an automatic semantic data integration:
  multi-agent framework approach, Semantic Web (2010) 107--134.

\bibitem{YAM2011}
D.~Ngo, Z.~Bellahsene, R.~Coletta, Yam++ - results for oaei 2011, in:
  Proceedings of the 6th International Workshop on Ontology Matching (OM2011).
  Bonn, Germany, 2011.

\bibitem{YAM2016}
D.~Ngo, Z.~Bellahsene, Overview of yam++—(not) yet another matcher for
  ontology alignment task, Journal of Web Semantics 41 (2016) 30--49.

\bibitem{CODI2010}
M.~Niepert, C.~Meilicke, H.~Stuckenschmidt, A probabilistic-logical framework
  for ontology matching, in: Proceedings of the 24th AAAI Conference on
  Artificial Intelligence (AAAI2010). Georgia, USA, 2010.

\bibitem{CODI2011}
J.~Huber, T.~Sztyler, J.~Noessner, et~al., Codi: combinatorial optimization for
  data integration - results for oaei 2011, in: Proceedings of the 6th
  International Workshop on Ontology Matching (OM2011). Bonn, Germany, 2011.

\bibitem{Kang2003}
J.~Kang, J.~F. Naughton, On schema matching with opaque column names and data
  values, in: Proceedings of the 2003 ACM SIGMOD international conference on
  Management of data (SIGMOD2003). California, USA, 2003.

\bibitem{KG2014}
X.~Dong, E.~Gabrilovich, G.~Heitz, et~al., Knowledge vault: a web-scale
  approach to probabilistic knowledge fusion, in: Proceedings of the 20th ACM
  SIGKDD International Conference on Knowledge Discovery and Data Mining
  (KDD2014). New York, USA, 2014.

\bibitem{Mikolov2013}
T.~Mikolov, I.~Sutskever, K.~Chen, et~al., Distributed representations of words
  and phrases and their compositionality, in: Proceedings of the 27th Annual
  Conference on Neural Information Processing Systems (NIPS2013). Nevada, USA,
  2013.

\bibitem{Bordes2013}
A.~Bordes, N.~Usunier, A.~Garcia-Duran, et~al., Translating embeddings for
  modeling multi-relational data, in: Proceedings of the 27th Annual Conference
  on Neural Information Processing Systems (NIPS2013). Nevada, USA, 2013.

\end{thebibliography}







\end{document}